# RetiWave-Mamba: A Dual-Stream Network for Retinal Disease Detection based on Multi-scale Context and Feature-Adaptive Mamba Projection

Cheng Cheng[1,2], Jin Hong[1,*]

1. School of Artificial Intelligence, Nanchang University, Nanchang 330031, China
2. School of Information Engineering, Nanchang University, Nanchang 330031, China

E-mail: 6105123080@email.ncu.edu.cn; hongjin@ncu.edu.cn;
* Correspondence should be addressed to Jin Hong

**Abstract:** Retinal diseases are a leading cause of irreversible vision impairment, making early and accurate diagnosis essential for effective treatment. Optical Coherence Tomography (OCT) serves as a critical imaging modality for this purpose, yet its automated analysis is hindered by inherent speckle noise, varying lesion scales, and subtle inter-class similarities. To address these challenges, we propose a novel framework, RetiWave-Mamba, which integrates spatial-frequency domain learning with state-of-the-art state space models. The framework utilizes Discrete Wavelet Transform (DWT) to decompose OCT images into low- and high-frequency streams, enabling decoupled processing of structural context and fine-grained details. For the low-frequency branch, we design a Multi-scale Contextual Localization Module (MCLM), which synergizes multi-scale dilation with spatial attention to expand the global receptive field and precisely localize lesion regions. For the high-frequency branch, we introduce an Attention-Guided High-Resolution Network (AG-HRNet) equipped with an intelligent gating mechanism to suppress noise propagation during multi-scale interactions. Furthermore, a Feature-Adaptive Mamba Projector (FAMP) is incorporated to form complementary channel-wise feature paths and adaptively reweight them using Mamba-generated gates. Extensive experiments on the OCT-C8 dataset demonstrate that our approach achieves a state-of-the-art (SOTA) classification accuracy of 98.38 ± 0.12%, surpassing existing methods. These results highlight the effectiveness of RetiWave-Mamba in identifying retinal pathologies and support its potential for computer-aided OCT image analysis.

## 1. Introduction

The eye is the primary organ for human perception, playing a key role in capturing and processing information from the external environment. However, visual impairment constitutes a significant and growing global health challenge. According to recent studies, retinal diseases such as diabetic retinopathy (DR) and age-related macular degeneration (AMD) remain the leading causes of irreversible blindness worldwide [1, 2]. Crucially, the clinical outcomes of these conditions are highly time-dependent; studies have shown that early diagnosis and timely intervention can prevent up to 95% of vision loss associated with conditions like diabetic retinopathy [3, 4]. Therefore, accurate and early identification of retinal lesions is important for preserving vision and improving patient outcomes [3, 4].

Currently, Optical Coherence Tomography (OCT) is the standard non-invasive method for retinal exams [5]. It uses low-coherence light to create high-resolution images of retinal structures, allowing doctors to clearly see abnormalities associated with various eye diseases [6]. However, manual interpretation of massive OCT data poses significant clinical challenges. The diagnostic process is labor-intensive and time-consuming, placing a heavy burden on ophthalmologists, particularly in regions with limited medical resources [7]. Furthermore, diagnosis relies heavily on the clinician's subjective experience, often leading to inter-observer variability and potential misdiagnosis [8]. These limitations underscore the urgent need for automated deep learning systems to achieve efficient and objective diagnosis.

In recent years, deep learning techniques have greatly improved medical image analysis, becoming the core technology of automated retinal disease diagnosis [7]. Among these, Convolutional Neural Networks (CNNs), such as ResNet and VGG, have emerged as the dominant architecture due to their powerful ability to extract hierarchical features [9, 10]. Beyond basic classification, CNNs have made significant contributions by automatically learning representative spatial patterns from complex retinal images [11]. To further enhance performance, researchers have incorporated advanced strategies, such as multi-scale feature fusion and frequency-domain analysis, enabling CNNs to effectively handle lesions of varying sizes while suppressing noise [12-15]. These multidimensional capabilities allow CNN-based models to achieve high diagnostic precision that effectively supports clinical decision-making, establishing them as the standard backbone for retinal disease analysis systems [16].

However, accurately diagnosing retinal diseases remains a major challenge due to the severe speckle noise [17] and subtle inter-class similarities inherent in OCT images, as illustrated in **Fig. 1**. Specifically, standard CNN backbones are often constrained by their fixed receptive fields, making it difficult to balance large-scale contextual understanding with precise lesion localization [16]. Moreover, regarding multi-scale interaction, most existing networks rely on simple summation or concatenation for feature fusion. Such simple fusion mechanisms directly transfer background speckle noise from low-resolution branches to high-resolution ones, thereby affecting the high-resolution features [12, 18, 19]. Additionally, current approaches face significant difficulties in processing fine-grained edge and texture features. Lacking an efficient mechanism to capture long-range dependencies, these models often struggle to distinguish subtle textural differences between morphologically similar lesions, such as CNV and DRUSEN [20].

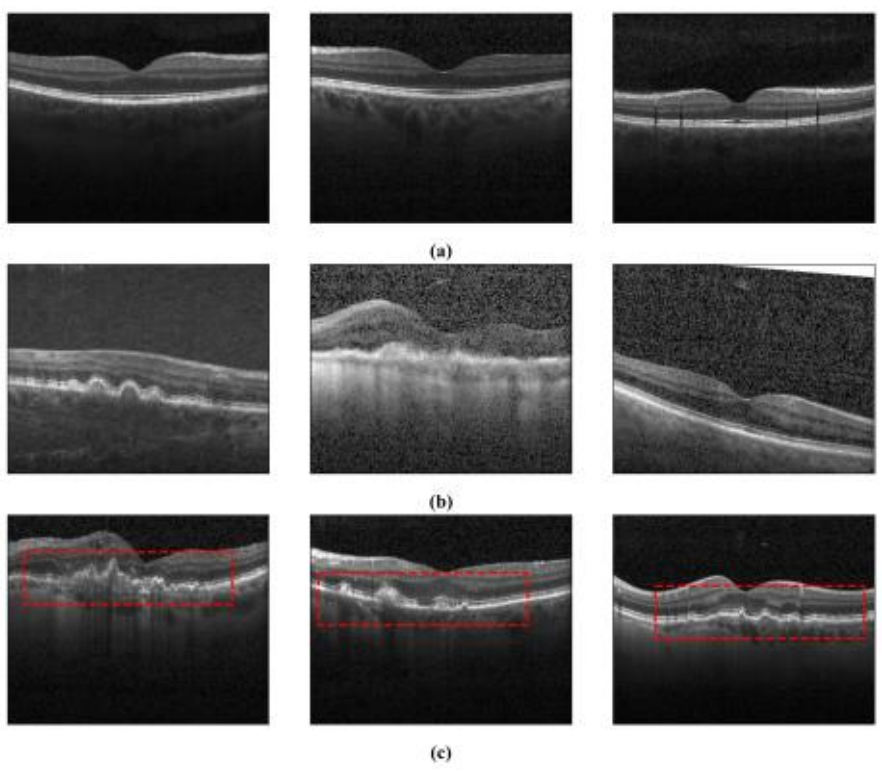

**Fig. 1** Visual illustration of the intrinsic challenges in automated OCT analysis. (a) Normal Retina: Representative images of healthy retinas exhibiting clear layer structures and distinct boundaries, serving as the diagnostic baseline. (b) Speckle Noise Interference: OCT scans severely corrupted by inherent speckle noise. Compared to the normal baseline in (a), the noise blurs structural boundaries and obscures fine-grained details. (c) Subtle Inter-class Similarities: Samples of CNV (left, middle) and DRUSEN (right). The red dashed boxes highlight the high morphological similarity in Retinal Pigment Epithelium (RPE) elevations, making these distinct categories difficult to differentiate without precise textural analysis.

More importantly, these challenges stem from the way information is represented in OCT images. Global retinal structures and large-scale lesion information are mainly represented by low-frequency components, while fine-grained edge and texture features are mainly contained in high-frequency components [14]. However, OCT speckle noise also produces strong high-frequency responses [17]. Therefore, removing too much high-frequency information may discard useful diagnostic details, whereas retaining all of it may increase noise. A conventional single-stream network processes global structural context, fine-grained edge and texture features, and speckle noise in the same feature space. This makes it difficult to preserve both global structural context and fine-grained features while suppressing noise. This trade-off motivates us to separate low-frequency and high-frequency information and process them in two specialized streams.

Based on this motivation, we propose RetiWave-Mamba, a dual-stream framework that processes low- and high-frequency information in separate streams using specialized modules. First, the Discrete Wavelet Transform (DWT) uses a fixed Haar wavelet to decompose the input image into low- and high-frequency components. The low-frequency stream focuses on global structural context and multi-scale lesion information. Its Multi-scale Contextual Localization Module (MCLM) captures contextual information from different receptive fields and enhances lesion-related features. The high-frequency stream processes fine-grained edge and texture features while reducing noise propagation. Its Attention-Guided High-Resolution Network (AG-HRNet) uses learned attention gates to control cross-resolution feature exchange, while the Feature-Adaptive Mamba Projector (FAMP) forms complementary channel-wise feature paths and adaptively reweights them within the DWT-derived high-frequency stream. In this framework, DWT performs the explicit wavelet-domain decomposition, while FAMP performs learned channel-wise feature partitioning, reweighting, and fusion within the DWT-derived high-frequency stream without introducing an additional frequency transform. The outputs of the two streams are then fused for retinal disease classification. The main contributions of this paper are as follows:

(i) We propose a hybrid dual-stream framework, RetiWave-Mamba, which combines spatial-domain and frequency-domain learning. By combining the MCLM, AG-HRNet, and FAMP modules, our approach captures both global structural contexts and detailed frequency features. This design leverages the strengths of different feature representations, significantly improving the model's performance in noisy environments.

(ii) We design a Multi-scale Contextual Localization Module (MCLM) for the low-frequency branch. By combining multi-scale dilation fusion with wavelet spatial attention, this module effectively expands the global receptive field while improving the model's ability to precisely localize lesion regions.

(iii) We develop an Attention-Guided High-Resolution Network (AG-HRNet). Unlike traditional architectures, AG-HRNet incorporates an intelligent gating mechanism during multi-scale interactions to replace simple summation, effectively reducing the propagation of background speckle noise and ensuring robust feature fusion.

(iv) We design a Feature-Adaptive Mamba Projector (FAMP) for the high-frequency stream. Rather than performing an additional frequency transform, FAMP forms two complementary channel-wise feature paths and uses Mamba-generated gates to adaptively reweight and fuse them, facilitating the integration of fine-grained edge and texture information.

(v) We evaluate our model on the publicly available OCT-C8 dataset, achieving a state-of-the-art (SOTA) accuracy of 98.38 ± 0.12%. This result demonstrates superior performance over existing methods and highlights the model’s discriminative capability, supporting its potential for computer-aided retinal disease classification.

## 2. Related work

### 2.1 Deep Learning for OCT Image Analysis

In recent years, deep learning has significantly advanced the field of OCT image analysis [7]. Convolutional Neural Networks (CNNs) have emerged as the dominant architecture for automated retinal diagnosis. Classic models, particularly ResNet-50 [9] and VGG [10], serve as robust backbones for hierarchical feature extraction. To further enhance diagnostic precision, researchers have introduced various structural optimizations to these baselines. For instance, Karthik et al. proposed "Edgen" blocks to replace standard residual connections, thereby improving the network's sensitivity to retinal boundaries [16]. Similarly, Fang et al. incorporated attention mechanisms to explicitly guide the model's focus toward critical lesion areas [13].

Beyond structural modifications to single networks, multi-scale feature learning and hybrid architectures have become critical strategies to accommodate the significant variation in lesion sizes. Researchers like Sotoudeh-Paima et al. [18] and Peng et al. [12] have developed multi-scale networks that aggregate features from different resolutions to enhance detection capabilities. To further capture features at various granularities, the OCTNet framework integrated an InceptionV3 backbone with a modified multi-scale spatial attention block, effectively extracting rich features from relevant lesion regions [21]. Moreover, the limitations of CNNs in modeling long-range dependencies have spurred the development of hybrid models. For instance, Laouarem et al. introduced HTC-Retina [20], and recent works like the SViT model have combined lightweight CNNs with Vision Transformers (ViTs) [22]. These approaches leverage the local inductive bias of CNNs and the global context awareness of Transformers to achieve high-precision classification.

Despite these advancements, current methods still face inherent limitations. Most multi-scale approaches rely on straightforward fusion mechanisms, such as element-wise summation or concatenation, which often fail to filter out speckle noise effectively, leading to the propagation of interference across scales [12, 18]. Furthermore, while hybrid Transformer models address the receptive field issue, they typically incur high computational complexity and may struggle to preserve the fine-grained high-frequency details required for detecting subtle retinal abnormalities. These challenges highlight the need for more efficient architectures capable of simultaneous noise suppression and precise feature representation. To address these limitations, we propose the RetiWave-Mamba framework, which synergizes spatial-frequency learning to effectively balance global structural understanding with fine-grained feature preservation.

### 2.2 Context Modeling and Attention Mechanisms

Accurate lesion localization requires a model to see the global context and focus on specific regions. To expand the receptive field without reducing resolution, dilated convolutions are widely used [23]. Architectures like CPFNet [24] and MSLI-Net [15] use multi-scale dilation strategies. They effectively aggregate contextual information and reduce the "gridding effect" found in stacked layers.

At the same time, attention mechanisms like SE [25], CBAM [26], and BAM [27] are used to suppress background noise. More recently, Coordinate Attention [28] has been proposed to capture long-range spatial dependencies with precise positional information, offering a lightweight alternative for medical imaging tasks. In retinal analysis, methods like LDCNN [13] utilize attention maps to highlight lesion areas. Similarly, Attention Gated Networks [29] explicitly learn to suppress irrelevant background regions in CT and ultrasound scans. These mechanisms help the model distinguish between useful features and background noise.

Related advances in remote-sensing detection have also shown that combining complementary information can improve feature discrimination under complex background interference. A spatial-temporal-spectral recurrent network integrates spatial context, temporal changes, and spectral differences for early-stage wildfire detection [30], while a recursive Transformer further fuses multi-type and multi-source information, including spectral, temporal, spatial, and land-cover cues, to reduce cloud and terrain interference [31]. More recently, dynamic context-adaptive fusion combines contextual statistics, multiple anomaly features, and spatiotemporal consistency constraints for 10-minute wildfire detection [32]. Although these methods focus on multi-temporal satellite observations for wildfire detection, they also show the value of adaptively combining complementary information under background interference.

Advanced remote-sensing image reconstruction methods further highlight the importance of efficient feature selection and attention-based context modeling. TTST dynamically selects the most relevant tokens and combines multi-scale feature aggregation with global context attention to reduce irrelevant information [33]. SpikeSR uses a spiking attention block to jointly model temporal, channel, and spatial information, improving feature representation while maintaining computational efficiency [34]. Unlike these remote-sensing detection and reconstruction methods, RetiWave-Mamba focuses on single OCT images and explicitly separates low- and high-frequency information for retinal disease classification.

However, treating context aggregation and attention as separate steps is not optimal. Standard CNN-based spatial attention often lacks a large receptive field. Meanwhile, pure dilated convolutions may aggregate noise along with the context. To overcome this, we design the Multi-scale Contextual Localization Module (MCLM), which establishes a unified mechanism that synergizes broad contextual understanding with noise-filtering capabilities to guide precise lesion localization.

### 2.3 Frequency-domain Learning in Medical Imaging

Traditional Convolutional Neural Networks primarily focus on extracting features in the spatial domain. However, frequency-domain learning offers a complementary perspective by decomposing images into distinct frequency components, which is highly effective for noise suppression and edge enhancement. The Discrete Wavelet Transform (DWT) has been widely adopted to separate images into low-frequency structural approximations and high-frequency textural details [35]. Inspired by this, architectures like WaveViT [36] have successfully integrated wavelet transforms into Vision Transformers to achieve lossless downsampling and multi-scale learning. Recent studies have further explored specialized feature streams [14, 15] and frequency-selection mechanisms [37] to improve the use of complementary feature information.

Among these methods, WaveNet-SF [14] is the closest spatial-frequency OCT classification framework to our work. It applies wavelet decomposition to construct low- and high-frequency streams and employs MSW-SA and HFFC to enhance lesion regions and compensate for high-frequency details, respectively. RetiWave-Mamba shares the general strategy of separately processing structural and detailed information but differs in the roles assigned to the two streams. MCLM models multi-scale contextual information in the low-frequency stream; AG-HRNet maintains multi-resolution representations in the high-frequency stream and controls cross-resolution exchange through learned gates; and FAMP constructs complementary channel-wise feature paths from the AG-HRNet outputs and uses Mamba-generated gates to reweight and fuse them. Thus, unlike the MSW-SA/HFFC enhancement pipeline in WaveNet-SF, RetiWave-Mamba coordinates contextual localization, gated multi-resolution representation, and Mamba-based feature projection after explicit wavelet decomposition.

Despite these advancements, effectively utilizing the decomposed high-frequency sub-bands remains a challenge, as they contain inextricably linked fine-grained textures and high-intensity speckle noise. Most existing hybrid methods typically process these components using standard convolutions and recombine them via simple element-wise summation or concatenation [14, 18]. These generic fusion strategies lack the spatial adaptivity to distinguish between valid edge details and background noise, often leading to the re-introduction of interference. Therefore, we introduce the Attention-Guided High-Resolution Network (AG-HRNet), utilizing an intelligent gating mechanism to selectively amplify diagnostic textures while suppressing noise propagation during multi-scale interactions.

### 2.4 State Space Models in Medical Imaging

Transformers have revolutionized long-sequence modeling with their exceptional global context awareness. However, their self-attention mechanism suffers from quadratic computational complexity, limiting efficiency on high-resolution data. To overcome this bottleneck, State Space Models (SSMs) have emerged as a promising alternative. Notably, building upon the Structured State Space sequence models [38], Mamba [39] utilizes a selective scan mechanism to capture long-range dependencies with linear time complexity. This breakthrough has rapidly extended to computer vision, where architectures like Vision Mamba [40] and VMamba [41] flatten images into sequences, demonstrating performance comparable to or exceeding traditional CNNs.

Mamba-based architectures have also been applied to medical image analysis and image restoration because of their efficient feature-modeling ability. For retinal disease detection, MRVM introduces a multi-directional selective mechanism to capture contextual information from OCT images [42]. Beyond OCT classification, several studies have combined Mamba with frequency information or task-specific feature representations. Frequency-Assisted Mamba for remote sensing image super-resolution [37] combines a Vision State Space Module with an FFT-based Frequency Selection Module to select informative spectral features during image reconstruction. VDMamba [43] constructs directional feature vectors through frequency-domain vector decomposition, while GraphMamba [44] models relationships among graph-structured instances in whole-slide images. PH-Mamba [45] introduces position-guided scanning and harmonized attention for image restoration.

RetiWave-Mamba differs from these methods in both its task and feature-processing strategy. Unlike Frequency-Assisted Mamba, which performs FFT-based frequency selection during image reconstruction, RetiWave-Mamba uses a fixed Haar DWT to explicitly decompose each OCT image into low- and high-frequency streams before feature extraction. FAMP does not perform another spectral transform. Instead, it forms two complementary channel-wise feature paths within the high-frequency stream and uses Mamba-generated gates to adaptively reweight and fuse them. Therefore, the contribution of RetiWave-Mamba lies in coordinating explicit frequency decomposition, low-frequency contextual localization, gated multi-resolution high-frequency representation, and feature-adaptive projection for OCT classification.

## 3. Method

### 3.1 Overall architecture

The overall framework of our proposed RetiWave-Mamba is illustrated in **Fig. 2**. The architecture is a hybrid dual-stream network comprising three key components: a Discrete Wavelet Transform (DWT) block, a low-frequency stream equipped with a Multi-scale Contextual Localization Module (MCLM), and a high-frequency stream integrating an Attention-Guided High-Resolution Network (AG-HRNet) with a Feature-Adaptive Mamba Projector (FAMP). This design enables the network to systematically process global structural context and fine-grained frequency details in parallel for robust retinal disease classification.

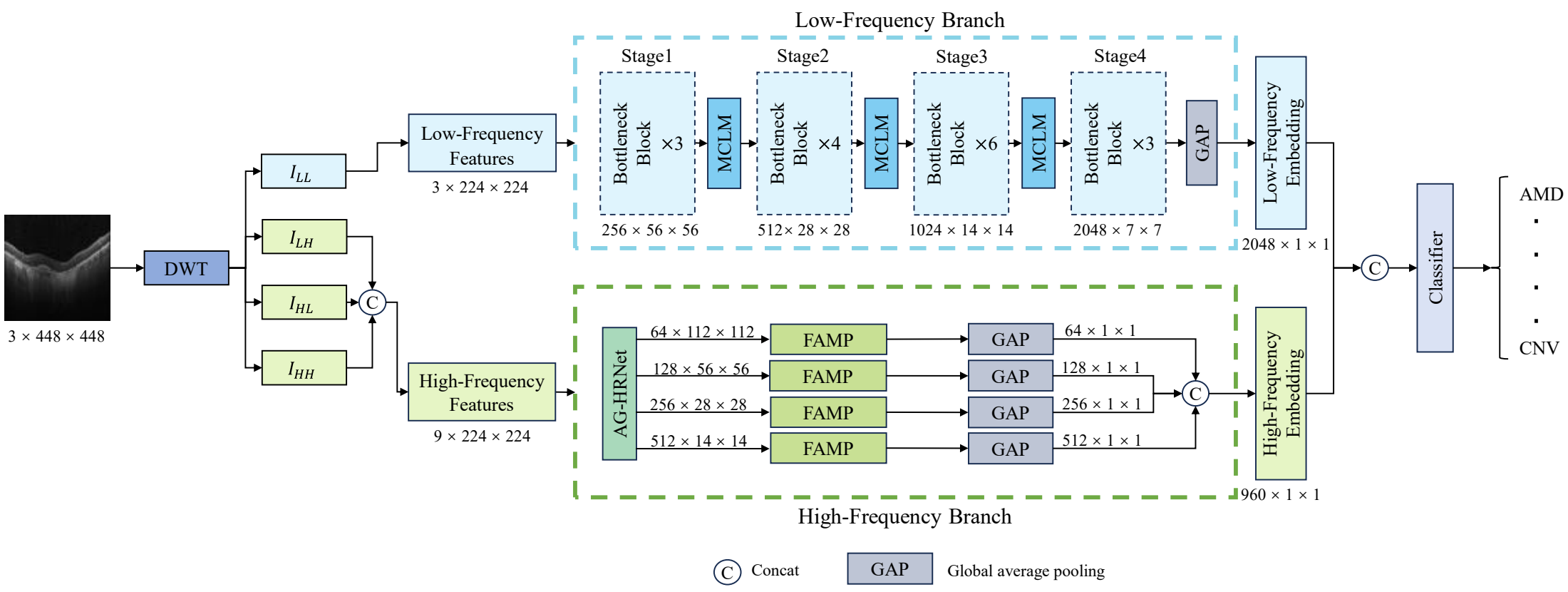

**Fig. 2** The overall architecture of the proposed RetiWave-Mamba framework.

Initially, the input OCT image is processed by the DWT block, which decomposes the image into one low-frequency (LF) component (the LL sub-band) and three high-frequency (HF) components (the LH, HL, and HH sub-bands). This initial decomposition effectively separates the image's foundational structures from its detailed textures, allowing each to be routed to a specialized processing stream.

Then, the LF component, which preserves the global structure and contextual information while inherently suppressing high-frequency noise, is directed into the low-frequency stream. This stream utilizes a ResNet-50 backbone structured into four stages (Stage 1-4) of Bottleneck Blocks. To enhance the backbone's receptive field and its ability to pinpoint lesion areas, our proposed MCLM is inserted after the first three stages. The feature map from the final stage then undergoes Global Average Pooling (GAP) to generate a compact low-frequency embedding vector.

Simultaneously, the three HF components (LH, HL, and HH) are concatenated along the channel dimension (creating a 9-channel input) and fed into the high-frequency stream to capture fine-grained edges and subtle textures critical for distinguishing similar lesions. This stream is initiated by the AG-HRNet, which maintains four parallel high-to-low resolution branches to preserve feature map fidelity. The four multi-scale outputs from AG-HRNet are then individually processed by FAMP modules, which construct complementary channel-wise feature paths and adaptively reweight them using Mamba-generated gates. Each projected feature map is subsequently processed by a separate GAP layer, and the resulting four feature vectors are concatenated to produce the final high-frequency embedding.

Finally, the low-frequency embedding and the high-frequency embedding are concatenated to form a comprehensive hybrid feature representation. This fused vector, which encapsulates complementary information from both the spatial-structural and frequency-textural domains, is fed into a final classifier to yield the ultimate diagnostic prediction.

### 3.2 Discrete Wavelet Transform (DWT)

The Discrete Wavelet Transform (DWT) is an effective method for analyzing the frequency domain information of images by decomposing them into localized frequency components. As illustrated in **Fig. 3**, this process is achieved by convolving the input image $I$ with separable low-pass ($L$) and high-pass ($H$) filters along its rows ($z_1$) and columns ($z_2$), followed by a 2x downsampling ($\downarrow_2$) at each stage. This decomposition results in four sub-bands, with the computation shown in Eq. (1) to (4):

$$I_{LL} = \left(I * L(z_1)\right) \downarrow_2 * L(z_2) \downarrow_2 \tag{1}$$

$$I_{LH} = \left(I * H(z_1)\right) \downarrow_2 * L(z_2) \downarrow_2 \tag{2}$$

$$I_{HL} = \left(I * L(z_1)\right) \downarrow_2 * H(z_2) \downarrow_2 \tag{3}$$

$$I_{HH} = \left(I * H(z_1)\right) \downarrow_2 * H(z_2) \downarrow_2 \tag{4}$$

where $*$ denotes the convolution operation, $I_{LL}$ represents the approximation component, and $I_{LH}$, $I_{HL}$, and $I_{HH}$ represent the detail components capturing horizontal, vertical, and diagonal information, respectively.

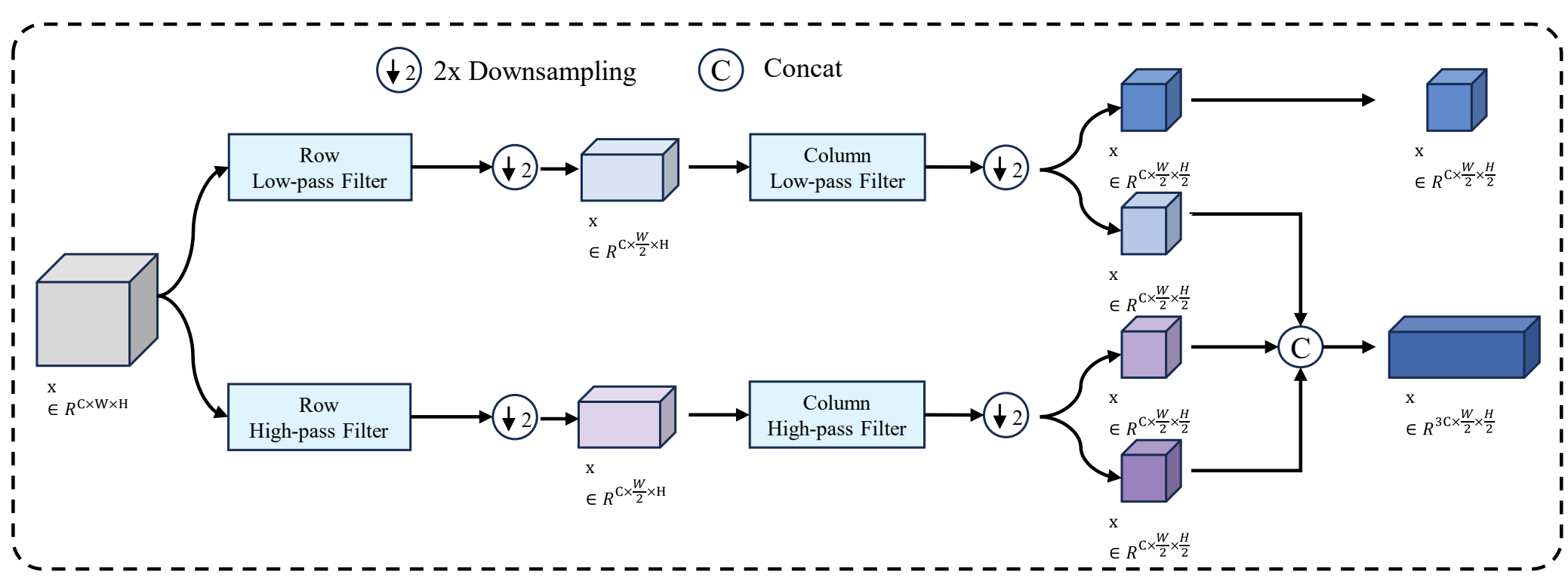

**Fig. 3** The decomposition process of the 2D Discrete Wavelet Transform (DWT).

In our framework, we selected the Haar wavelet for its computational simplicity and compatibility with the dual-stream design. Its two-tap filters compute local averages and differences, providing a spatially localized representation of intensity variations. The approximation component and all three detail components are retained and processed by the corresponding streams, allowing the subsequent networks to learn structural and fine-detail features without discarding any subband at the decomposition stage. Higher-order Daubechies, Symlets, and Coiflets have different filter properties [46] and may yield different representations of retinal structures and textures. Our choice of Haar prioritizes a simple, low-cost decomposition for separating these complementary types of information. The DWT module serves as the foundational processing step to bifurcate the image representation. The Low-Frequency (LF) Component, $I_{LL}$, which encapsulates the stable, denoised structural context of the retina, is isolated and directed as the input to our low-frequency stream.

Conversely, the three detail sub-bands ($I_{LH}$, $I_{HL}$, and $I_{HH}$) are designated as the High-Frequency (HF) Component. To form a comprehensive high-frequency representation, our model employs a key technique: we concatenate the detail sub-bands along the channel dimension. This approach is chosen over summation, as summation would obscure valuable directional information. For a 3-channel input image, this concatenation operation creates a 9-channel feature map. This resulting tensor preserves the distinct horizontal, vertical, and diagonal edge details, serving as a rich, multi-channel input for the specialized high-frequency stream.

### 3.3 Multi-scale Contextual Localization Module (MCLM)

As outlined in Section 3.1, the MCLM is the core innovative module we designed for the low-frequency stream, embedded after the first three stages of the ResNet-50 backbone. Its primary objective is to address the limited receptive field of standard CNNs and enhance the model's ability to precisely localize lesion regions against a noisy background. This placement allows MCLM to capture multi-scale context and refine lesion-related features while the spatial resolution remains sufficient for detailed feature processing. Applying MCLM after the first three stages therefore enables progressive feature refinement before further downsampling and provides enriched representations for subsequent stages. In contrast, the fourth stage has the lowest spatial resolution and the largest number of channels, and its output is directly passed to global average pooling for final feature aggregation. Adding another MCLM at this stage would increase the parameter and computational costs while offering limited additional spatial information. Therefore, we selected the first three stages as a practical compromise between multi-level feature refinement, spatial-detail preservation, and model complexity.

The detailed architecture of the MCLM is illustrated in **Fig. 4**. It consists of two cascaded sub-units: (a) the Multi-scale Dilation Fusion (MDF) unit, which captures wide-range contextual information; and (b) the Multi-Scale Wavelet Spatial Attention (MSW_SA) unit, which refines the features and enhances the salience of lesion regions.

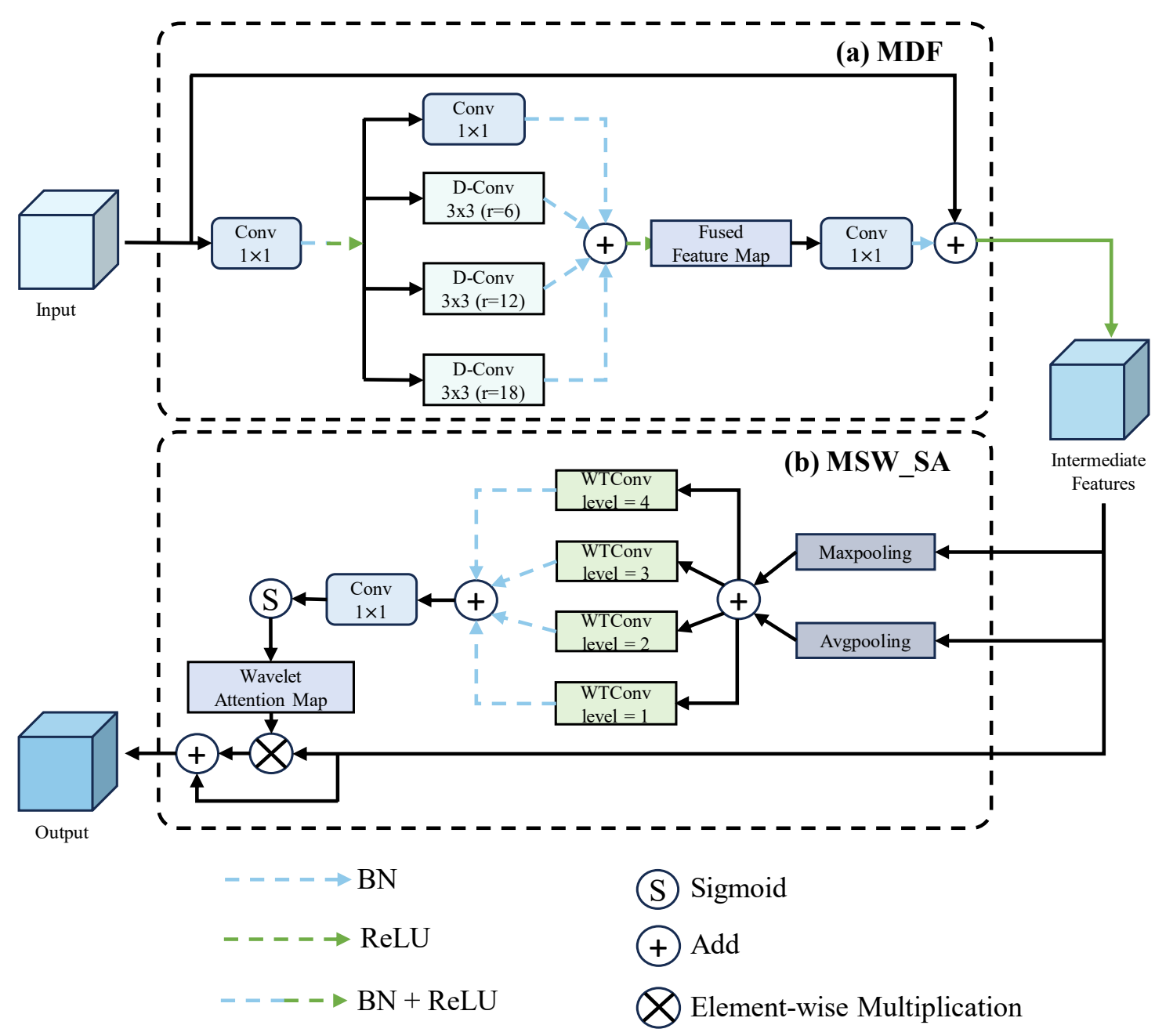

**Fig. 4** The architecture of the Multi-scale Contextual Localization Module (MCLM). The module consists of two components: (a) the Multi-scale Dilation Fusion (MDF) unit, which captures multi-scale context, and (b) the Multi-Scale Wavelet Spatial Attention (MSW_SA) unit, which refines the features to localize lesion regions.

### *3.3.1 Multi-scale Dilation Fusion (MDF)*

Standard CNNs often struggle to balance large-scale contextual understanding with the preservation of spatial resolution, as their fixed receptive fields limit their ability to model long-range dependencies. While dilated convolutions can expand the receptive field without downsampling, stacking them can lead to gridding artifacts, and simple parallel structures may not effectively fuse features from different scales.

To address this, we introduce the Multi-scale Dilation Fusion (MDF) unit, which is designed to efficiently capture and fuse multi-scale features. The architecture of this unit is shown in **Fig. 4(a)**.

The MDF unit is designed to efficiently capture and fuse multi-scale features. Let $X$ be the input feature map. First, a projection function is applied to reduce the channel dimensionality. This operation comprises a 1×1 convolution, followed by batch normalization and the ReLU activation function. This can be expressed by Eq. (5):

$$X' = \sigma\left(\mathcal{B}\left(W_{proj} * X\right)\right) \tag{5}$$

where $\mathcal{B}(\cdot)$ denotes batch normalization, $\sigma(\cdot)$ denotes the ReLU activation function, and $W_{proj}$ is the projection convolution kernel.

Then, a set of dilation rates $R = \{6,12,18\}$ is employed to cover larger receptive fields. The feature $X'$ is processed by one standard convolution branch and three dilated convolution branches, which are subsequently fused via element-wise summation. This process is formulated in Eq. (6):

$$F_r = \sigma\left(\mathcal{B}(W_1 * X') + \sum_{r \in R} \mathcal{B}\left(W_{3,r} *_r X'\right)\right) \tag{6}$$

where $W_1$ represents the kernel for the standard branch, $W_{3,r}$ denotes the kernels with dilation rate $r$, and $*_r$ indicates the dilated convolution.

Finally, the feature $F_r$ is projected back to the original dimension using a 1×1 convolution kernel denoted as $W_{out}$. Subsequently, it is fused with the original input $X$ via a residual connection to obtain the output feature $F$, as shown in Eq. (7):

$$F = \sigma(\mathcal{B}(W_{out} * F_r) + X) \tag{7}$$

By fusing these parallel dilated branches, the MDF unit effectively expands the receptive field to capture large-scale dependencies. Simultaneously, the residual connection and the 1×1 convolutional branch ensure that fine-grained local details are preserved, providing a contextually rich feature map $F$ for the subsequent attention unit.

*3.3.2 Multi-Scale Wavelet Spatial Attention (MSW_SA)*

While the MDF unit effectively captures multi-scale context, the resulting intermediate feature map $F$ can still contain irrelevant background noise. Standard spatial attention mechanisms often use large-kernel convolutions to perceive spatial context, which can be computationally expensive and may not effectively distinguish lesion signals from noise, especially in OCT images.

To address the background noise in the intermediate feature map $F$, the MSW_SA unit applies a targeted spatial attention weighting. The core of MSW_SA is the generation of a refined spatial attention map, $M_S$. The process begins by aggregating channel information. The input feature $F$ undergoes max pooling and average pooling, and the results are combined via element-wise summation. This aggregated feature is then fed into the WTConv module, which consists of four parallel branches. The outputs of these four branches are normalized by batch normalization and summed together. Subsequently, the result is processed by a 1×1 convolution and the sigmoid activation function to generate the final map $M_S$. The calculation process is as follows:

$$M_S = \delta\left(W_{MSW_SA} * \sum_{k=1}^{4} \mathcal{B}\left(W_{wt,k}\left(P_{max}(F) + P_{avg}(F)\right)\right)\right) \tag{8}$$

where $\delta(\cdot)$ represents the sigmoid activation function, $W_{MSW_SA}$ represents the 1×1 convolution kernel for fusion, $W_{wt,k}$ represents the wavelet convolution kernel at level $k$ (ranging from 1 to 4), and $P_{max}(\cdot)$ and $P_{avg}(\cdot)$ represent max pooling and average pooling, respectively.

Finally, this wavelet attention map $M_S$ is applied to the original intermediate feature map $F$ via element-wise multiplication, which is then combined with a residual connection to produce the final output of the MCLM module, $y$. This process, which allows the model to enhance lesion features while preserving the original contextual information, is shown in Eq. (9):

$$y = F + (F \otimes M_S) \tag{9}$$

where $\otimes$ represents element-wise multiplication.

This cascaded design allows the global structural information, first captured by the MDF, to be intelligently re-weighted by the wavelet-based spatial attention. This effectively guides the model to focus on salient lesion structures while reducing the susceptibility to the complex background noise that often interferes with traditional spatial attention mechanisms.

### 3.4 Attention-Guided High-Resolution Network (AG-HRNet)

The Attention-Guided High-Resolution Network, or AG-HRNet, is the core component of the high-frequency stream, designed to process fine-grained edge and texture details. Its primary function is to maintain high-resolution feature representations in parallel across multiple scales, as illustrated in the architecture diagram in **Fig. 5**. The network's operation begins by establishing this parallel, multi-stage architecture. It starts with a high-resolution branch and progressively adds lower-resolution branches through four stages. The creation of these new branches is handled by Stage Transitions, which are standard strided convolutions. This parallel structure is the key to preserving the high-resolution features essential for processing fine-grained details.

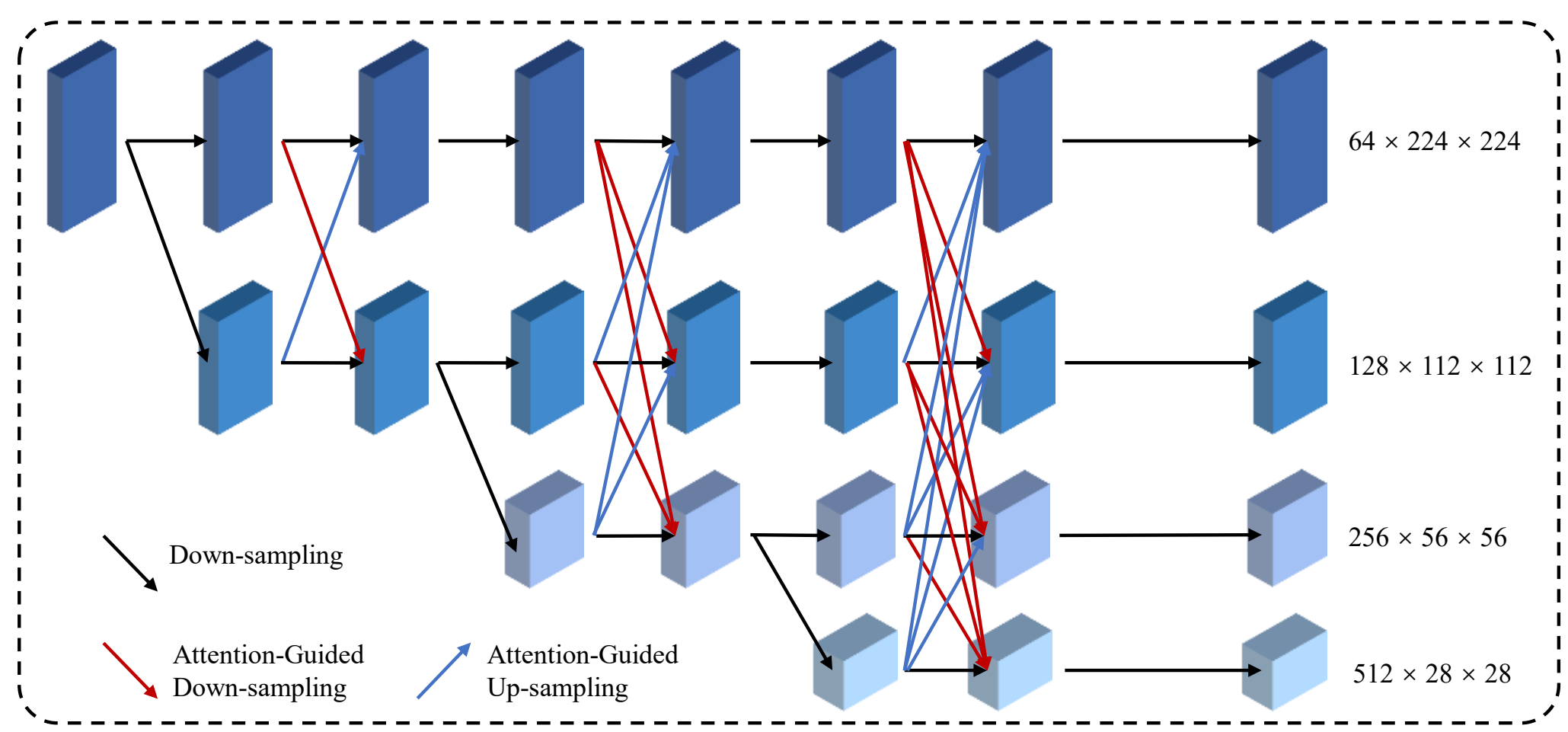

**Fig. 5** The architecture of the Attention-Guided High-Resolution Network (AG-HRNet). The module operates on parallel branches. Diagram Key: Black arrows represent Standard Down-sampling for stage transitions. Red arrows represent Attention-Guided Down-sampling, and blue arrows represent Attention-Guided Up-sampling. All colored arrows indicate the use of the Attention-Guided Fusion (AGF) mechanism.

The core innovation of this module lies in how it facilitates the crucial interaction between these parallel branches. Instead of the simple summation used in traditional multi-scale architectures such as HRNet [47], which can propagate noise, the AG-HRNet employs a novel Attention-Guided Fusion mechanism, or AGF. This mechanism operates as the exclusive pathway for all cross-resolution feature exchanges.

The AGF mechanism works as a smart gate. Let $F_h$ be the feature map from the target branch, and $F_l'$ be the feature map from the source branch. After adjusting $F_l'$ to the same size as $F_h$, an attention gate $G$ is generated from the source features. As shown in Eq. (10), a 1×1 convolution denoted as $W_{gate}$ is used, followed by batch normalization and a sigmoid activation function.

$$G = \delta\left(\mathcal{B}(W_{gate} * F_l')\right) \tag{10}$$

Then, this gate $G$ is applied to the incoming features $F_l'$ via element-wise multiplication. This step operates as a selective filter, re-weighting the incoming features to amplify useful information and suppress noise. The final fusion is then completed by adding this filtered map to the target branch $F_h$ via a residual connection to obtain the fused feature $y_{fused}$. This is shown in Eq. (11):

$$y_{fused} = F_h + (F_l' \otimes G) \tag{11}$$

Through this gate-and-add operation, the AG-HRNet ensures that only high-confidence, relevant textural information is propagated between branches. This specific mechanism is the key to effectively reducing the speckle noise contamination that often plagues simple summation, ensuring the final high-resolution features remain clear and discriminative. The module outputs four multi-scale, noise-suppressed feature maps, which serve as ideal inputs for the subsequent FAMP modules.

### 3.5 Feature-Adaptive Mamba Projector (FAMP)

The Feature-Adaptive Mamba Projector (FAMP) processes each of the four multi-scale feature maps produced by AG-HRNet. The LL, LH, HL, and HH subbands are generated by DWT before feature extraction; therefore, FAMP does not perform a second frequency decomposition. For each AG-HRNet output, FAMP uses a learned channel mask and its complement to form two complementary feature paths. A joint descriptor of the two paths is then modeled by Mamba to generate path-specific weights, after which the reweighted paths are fused through a residual connection. The detailed operation and internal structure of FAMP are illustrated in **Fig. 6**.

The operation begins with a channel-gating mechanism. The input feature map $X$ is first globally pooled to summarize the response of each channel and generate a channel-wise attention mask $M$. As shown in Eq. (12), the pooled descriptor is processed by a learnable 1×1 channel-gating transform ($W_g$) followed by a sigmoid activation function:

$$M = \delta\left(W_g * P_{avg}(X)\right) \tag{12}$$

The learned mask is derived from the global channel responses of the input $X$ and, together with its complement, forms two softly partitioned feature paths, $X_A$ and $X_B$. The two paths are calculated by element-wise multiplication, as shown in Eqs. (13) and (14). Because the masks are complementary, information attenuated in one path is retained in the other, providing two complementary views for subsequent adaptive fusion:

$$X_A = X \otimes M \tag{13}$$

$$X_B = X \otimes (1 - M) \tag{14}$$

Although FAMP also uses global pooling to obtain channel information, its operation differs from that of an SE block. An SE block generates a single channel-attention vector and directly reweights the original feature map. In contrast, FAMP uses the learned mask $M$ and its complement $(1 - M)$ to form two softly partitioned feature paths. Their joint descriptor is then processed by Mamba to generate path-specific weights, after which the two reweighted paths are fused through a residual connection. Therefore, FAMP uses channel information to construct and coordinate complementary feature paths rather than only recalibrating the channels of a single feature map.

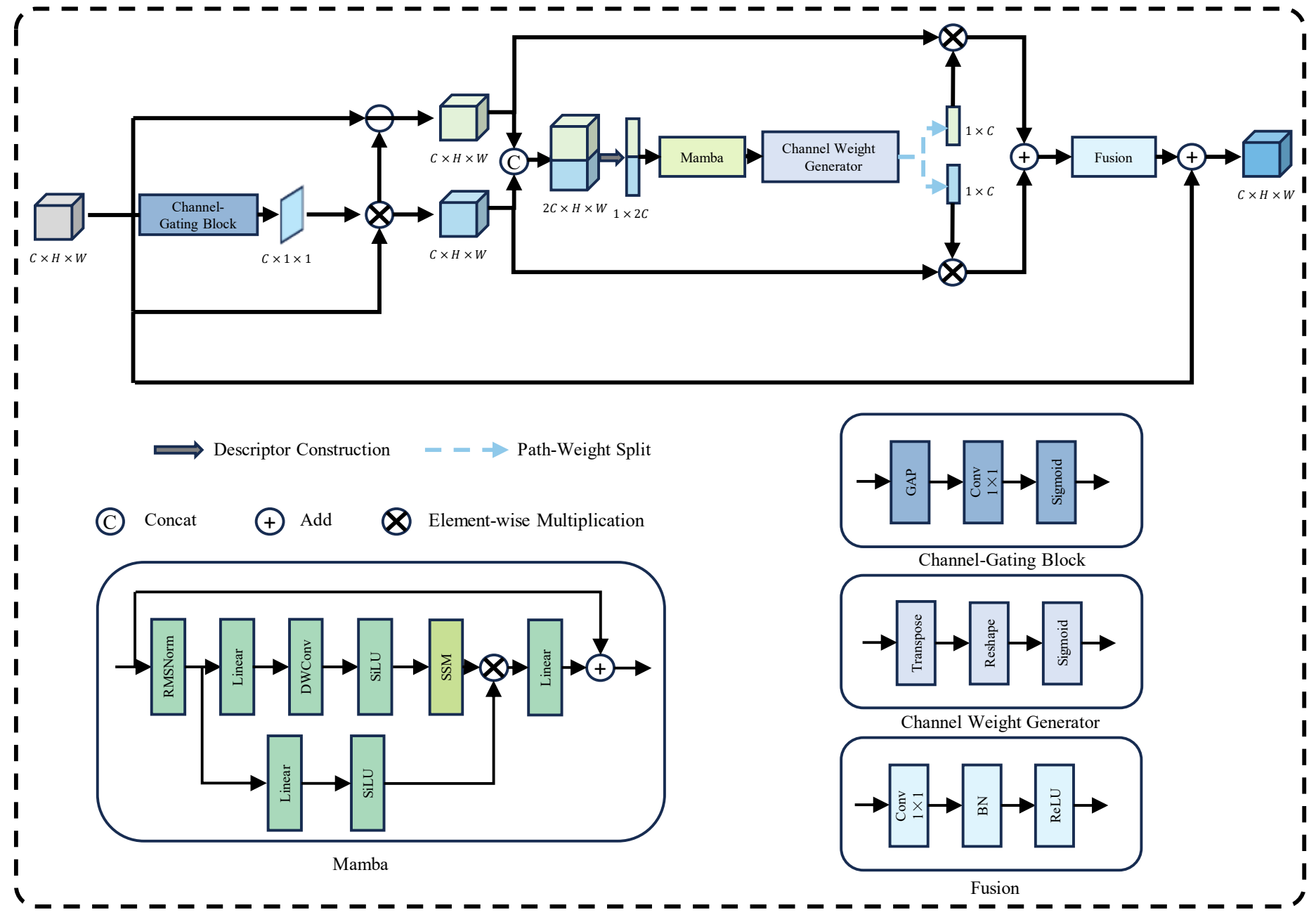

**Fig. 6** Architecture of the Feature-Adaptive Mamba Projector (FAMP). The Channel-Gating Block generates a channel-wise mask $M$ from the input feature map $X$. The mask $M$ and its complement $1 - M$ softly partition $X$ into two complementary feature paths, $X_A$ and $X_B$. After concatenation, global pooling, flattening, and normalization are applied to construct a joint channel descriptor. The descriptor is then processed by the Mamba block and the Channel Weight Generator to obtain the path-specific weights $W_A$ and $W_B$. The two reweighted paths are combined by the Fusion block and added to the original input through a residual connection to produce the output $Y$. The lower panels show the internal structures of the Channel-Gating Block, Mamba block, Channel Weight Generator, and Fusion block.

After the soft partition, $X_A$ and $X_B$ are concatenated along the channel dimension and converted into a joint channel descriptor through global pooling, flattening, and normalization. This operation summarizes the complementary information from both paths before Mamba-based gating and produces the representation $X_{seq}$. This is shown in Eq. (15):

$$X_{seq} = \mathcal{T}\left(\mathcal{C}(X_A, X_B)\right) \tag{15}$$

where $\mathcal{C}(\cdot)$ represents the concatenation operation along the channel dimension and $\mathcal{T}(\cdot)$ represents the transformation function comprising global pooling, flattening, and normalization.

The resulting descriptor $X_{seq}$ is then processed by the Mamba module [38]. The Mamba output is passed to a weight generator, which projects it back to the channel dimension. A sigmoid function is then applied to generate the attention tensor $W$, as described in Eq. (16):

$$W = \delta\left(\mathcal{P}\left(\mathcal{M}\left(X_{seq}\right)\right)\right) \tag{16}$$

where $\mathcal{P}(\cdot)$ represents the projection function and $\mathcal{M}(\cdot)$ represents the Mamba module operation.

The attention tensor is split into two path-specific gating vectors, $W_A$ and $W_B$. During fusion, $W_A$ reweights $X_A$, while $W_B$ reweights $X_B$. The two modulated paths are added and passed through the Fusion block $F_{fusion}(\cdot)$. This block consists of a 1×1 convolution ($W_{fusion}$), batch normalization, and ReLU activation, and is defined as:

$$F_{fusion}(Z) = \sigma\left(\mathcal{B}\left(W_{fusion} * Z\right)\right) \tag{17}$$

Finally, the output of this block is added to the original input $X$ via a residual connection. This produces the final output $Y$. This process is expressed in Eq. (18):

$$Y = X + F_{fusion}\big((X_A \otimes W_A) + (X_B \otimes W_B)\big) \tag{18}$$

Through this complementary soft partition and path-specific reweighting, FAMP adaptively balances channel responses while retaining information that might otherwise be suppressed by a single attention path.

## 4. Experiments

### 4.1 Dataset

To evaluate the performance of the proposed RetiWave-Mamba framework, we use the publicly available OCT-C8 dataset [48]. This dataset contains a total of 24,000 optical coherence tomography (OCT) images, evenly distributed across eight categories: Age-related Macular Degeneration (AMD), Choroidal Neovascularization (CNV), Central Serous Retinopathy (CSR), Diabetic Macular Edema (DME), Diabetic Retinopathy (DR), DRUSEN, Macular Hole (MH), and Normal, with 3,000 images per category. While the dataset is officially divided into training, validation, and test sets, we modified this split to increase the number of training samples. Following the protocol used in [15], we combined the original training set (18,400 images) and validation set (2,800 images) to create a new, larger training set of 21,200 images (2,650 per class). The official test set, containing 2,800 images (350 per class), remains unchanged and is used exclusively for evaluating the final model performance. Model checkpoints were selected according to the highest training accuracy rather than test-set performance.

### 4.2 Evaluation metrics

To quantitatively evaluate the classification performance of our proposed model, we use four standard metrics: Accuracy (ACC), Precision, Sensitivity (also known as Recall), and F1-score. These metrics are calculated based on the confusion matrix, using the counts of True Positive (TP), True Negative (TN), False Positive (FP), and False Negative (FN). The formulas for these metrics are as follows:

$$Accuracy = \frac{TP + TN}{TP + TN + FP + FN} \tag{19}$$

$$Precision = \frac{TP}{TP + FP} \tag{20}$$

$$Sensitivity = \frac{TP}{TP + FN} \tag{21}$$

$$F1 - score = \frac{2 \times Precision \times Sensitivity}{Precision + Sensitivity} \tag{22}$$

Here, Precision reflects the accuracy of the model's positive predictions, while Sensitivity reflects the model's ability to identify all true positive samples. The F1-score provides a balanced measure between Precision and Sensitivity.

### 4.3 Implementation details

The training and testing of this experiment were conducted on a single NVIDIA GeForce RTX 4090 GPU with 24GB of video memory. In the data preprocessing stage, we uniformly resized the input images to 448×448 pixels and normalized the images using pre-calculated mean and standard deviation. To enhance model generalization, data augmentation strategies such as random rotation and horizontal flipping were applied. During the training process, the loss function was chosen to be Weighted Cross-Entropy Loss to address class imbalance, and the model was optimized using the Adam optimizer, where the optimizer parameters were set to $\beta_1$=0.9 and $\beta_2$=0.999. The weight decay parameter was set to $1\times10^{-4}$. Additionally, the learning rate followed a cosine annealing schedule with a 5-epoch warm-up, starting at $2\times10^{-4}$, the batch size was set to 32, and the training duration was 60 epochs. To improve training efficiency and reduce memory consumption, the Automatic Mixed Precision (AMP) technique was employed.

For each reimplemented model, six runs were conducted with different random seeds under the same experimental protocol. Classification results are reported as mean ± standard deviation across runs. For the OCT-C8 comparison in **Tab. 1**, two-sided paired t-tests were performed on accuracy values matched by random seed between RetiWave-Mamba and each reimplemented baseline. The reported p-values are unadjusted, with a significance threshold of 0.05. Results taken from published studies are presented as originally reported and are excluded from these statistical tests.

### 4.4 Performance of the Proposed Method

In this section, we comprehensively evaluate the classification performance of the proposed RetiWave-Mamba framework on the OCT-C8 dataset. To validate the superiority of our method, we conducted a systematic comparison with several representative deep learning architectures. These include classic Convolutional Neural Networks (CNNs) such as ResNet-50 [9], VGG16 [10], GoogLeNet [49], DenseNet121 [50], and InceptionV3 [51]. Furthermore, we compared our approach with state-of-the-art (SOTA) models designed for efficient image analysis and medical imaging, including EfficientNet-B3 [52], ConvNeXt V2 [53], and the vision transformer-based Swin-Tiny [54]. To ensure a comprehensive evaluation, we also included specialized retinal classification networks. We selected FPN-ResNet50 [12] and FPN-DenseNet121 [18] to represent multi-scale perception methods. We also included ResNet-EdgeEn [16] for its focus on boundary enhancement. For hybrid architectures, we compared against Swin-Poly Transformer [55] and HTC-Retina [20]. Additionally, we included specialized retinal disease classification networks such as MSLI-Net [15] and WaveNet-SF [14] for a comprehensive evaluation.

For ResNet-EdgeEn, Swin-Poly Transformer, FPN-ResNet50, FPN-DenseNet121, and HTC-Retina, we directly adopted the results reported in their corresponding studies because these methods were evaluated using the same dataset, data partition, and classification task. The remaining comparison models were reimplemented under our experimental conditions.

**Tab. 1** presents a comprehensive comparison of RetiWave-Mamba with several representative deep learning models on the OCT-C8 dataset in terms of classification performance, model complexity, and inference efficiency. Our method achieved the highest mean accuracy of 98.38 ± 0.12%. Among the evaluated baseline models, InceptionV3 achieved the highest mean accuracy of 98.16 ± 0.21%, followed by DenseNet121 and Swin-Tiny, which achieved 97.90 ± 0.13% and 97.88 ± 0.11%, respectively. GoogLeNet, EfficientNet-B3, VGG16, and WaveNet-SF achieved mean accuracies of 97.77 ± 0.11%, 97.74 ± 0.16%, 97.74 ± 0.10%, and 97.73 ± 0.13%, respectively. In terms of precision, sensitivity, and F1-score, RetiWave-Mamba also achieved the best results, reaching 98.39 ± 0.11%, 98.38 ± 0.12%, and 98.38 ± 0.12%, respectively. Compared with the standard ResNet50 baseline, RetiWave-Mamba achieved an improvement of 0.79 percentage points in mean accuracy. Paired t-tests confirmed that the improvements over all reimplemented comparison models other than InceptionV3 were statistically significant ($p < 0.05$). These results confirm the effectiveness of RetiWave-Mamba across multiple evaluation metrics and demonstrate that the integration of the dual-stream wavelet architecture with Mamba-based long-range dependency modeling improves the extraction of both global structural information and fine-grained lesion details.

**Fig. 7** illustrates the training dynamics of RetiWave-Mamba. The left plot shows the accuracy curve, while the right plot displays the loss curve over 60 epochs. The curves indicate rapid convergence during the early stages of training and stable optimization in the later epochs, without obvious oscillation. Because the original validation set was merged into the training set, these curves are used only to describe the optimization process and are not interpreted as evidence that overfitting is absent.

To further analyze the model's discriminative capability for specific retinal diseases, **Fig. 8** displays the confusion matrix on the test set. The model exhibits exceptional performance in identifying distinct categories such as AMD, CNV, and Normal, achieving near-perfect classification accuracy. Notably, even for categories with high inter-class similarity, such as DRUSEN and CNV, our model maintains a low misclassification rate. This indicates that the high-frequency stream in RetiWave-Mamba effectively captures the subtle textural discrepancies required to distinguish morphologically similar lesions, thereby proving its potential for reliable clinical diagnosis.

**Tab. 1** Performance comparison of the OCT-C8 dataset

| Method | Accuracy | Precision | Sensitivity | F1-Score | P-value | Params (M) | FLOPs (G) | Inference time (ms) |
|---|---|---|---|---|---|---|---|---|
| VGG16[10] | 97.74 ± 0.10 | 97.75 ± 0.10 | 97.74 ± 0.10 | 97.74 ± 0.10 | $4.21 \times 10^{-4}$ | 134.29 | 61.51 | **3.06** |
| GoogLeNet[49] | 97.77 ± 0.11 | 97.78 ± 0.11 | 97.77 ± 0.11 | 97.77 ± 0.11 | $2.18 \times 10^{-4}$ | 5.61 | 6.04 | 5.30 |
| ResNet50[9] | 97.59 ± 0.14 | 97.60 ± 0.13 | 97.59 ± 0.14 | 97.59 ± 0.14 | $3.49 \times 10^{-4}$ | 23.52 | 16.53 | 3.97 |
| InceptionV3[51] | 98.16 ± 0.21 | 98.17 ± 0.20 | 98.16 ± 0.21 | 98.16 ± 0.21 | $9.99 \times 10^{-2}$ | 21.80 | 13.25 | 7.86 |
| DenseNet121[50] | 97.90 ± 0.13 | 97.91 ± 0.14 | 97.90 ± 0.13 | 97.90 ± 0.13 | $5.50 \times 10^{-4}$ | 6.96 | 11.58 | 11.66 |
| EfficientNet-B3[52] | 97.74 ± 0.16 | 97.76 ± 0.16 | 97.74 ± 0.16 | 97.74 ± 0.16 | $1.49 \times 10^{-3}$ | 10.71 | **4.07** | 8.51 |
| Swin-Tiny[54] | 97.88 ± 0.11 | 97.88 ± 0.11 | 97.88 ± 0.11 | 97.87 ± 0.11 | $1.00 \times 10^{-5}$ | 27.53 | 17.48 | 7.19 |
| ConvNeXt V2[53] | 97.20 ± 0.23 | 97.22 ± 0.22 | 97.20 ± 0.23 | 97.20 ± 0.23 | $4.03 \times 10^{-5}$ | 27.87 | 17.82 | 6.30 |
| ResNet-EdgeEn[16] | 92.40 | 93.00 | 92.00 | 92.00 | - | 21.29 | 14.73 | 4.35 |
| Swin-Poly Transformer[55] | 97.11 | 97.13 | 97.11 | 97.10 | - | 27.53 | 17.48 | 7.27 |
| FPN-ResNet50[12] | 97.14 | 97.15 | 97.14 | 97.14 | - | 31.07 | 97.75 | 5.02 |
| FPN-DenseNet121[18] | 97.22 | 97.23 | 97.22 | 97.21 | - | 14.25 | 92.76 | 11.62 |
| HTC-Retina[20] | 97.00 | 97.04 | 97.00 | 97.01 | - | **3.82** | 15.25 | 5.66 |
| MSLI-Net[15] | 97.43 ± 0.21 | 97.44 ± 0.20 | 97.43 ± 0.21 | 97.43 ± 0.21 | $3.25 \times 10^{-4}$ | 86.29 | 66.91 | 14.85 |
| WaveNet-SF[14] | 97.73 ± 0.13 | 97.73 ± 0.12 | 97.73 ± 0.13 | 97.73 ± 0.13 | $7.42 \times 10^{-5}$ | 27.84 | 6.66 | 12.62 |
| **RetiWave-Mamba (ours)** | **98.38 ± 0.12** | **98.39 ± 0.11** | **98.38 ± 0.12** | **98.38 ± 0.12** | **-** | 65.92 | 29.48 | 19.99 |

In addition, to comprehensively evaluate the efficiency of each model, we compare three metrics: the number of parameters, floating-point operations (FLOPs), and inference time per image. The number of parameters reflects model size and storage requirements, FLOPs estimate the computational cost of a single forward pass, and inference time indicates actual execution speed in the same hardware environment.

**Tab. 1** also reports the model complexity and inference efficiency of all evaluated methods. RetiWave-Mamba contains 65.92 M parameters and requires 29.48 G FLOPs, with an average inference time of 19.99 ms per image. Its parameter count and computational cost are higher than those of most conventional backbones because the dual-stream architecture combines high-resolution feature processing with Mamba-based projection. Compared with VGG16, RetiWave-Mamba reduces the number of parameters and FLOPs by 50.91% and 52.08%, respectively, while improving the mean accuracy from 97.74 ± 0.10% to 98.38 ± 0.12%. Compared with FPN-ResNet50 and FPN-DenseNet121, RetiWave-Mamba uses more parameters but requires considerably fewer FLOPs (29.48 G versus 97.75 G and 92.76 G) and achieves a higher mean accuracy (98.38 ± 0.12% versus the reported accuracies of 97.14% and 97.22%). Compared with MSLI-Net, it reduces the number of parameters from 86.29 M to 65.92 M and the FLOPs from 66.91 G to 29.48 G, while improving the mean accuracy from 97.43 ± 0.21% to 98.38 ± 0.12%. However, its inference time increases from 14.85 ms to 19.99 ms. Although its inference time is the highest among the evaluated models, it remains below 20 ms per image under the tested setting. Together with the results in **Tab. 1**, these comparisons show that RetiWave-Mamba achieves higher classification performance with some additional computational cost.

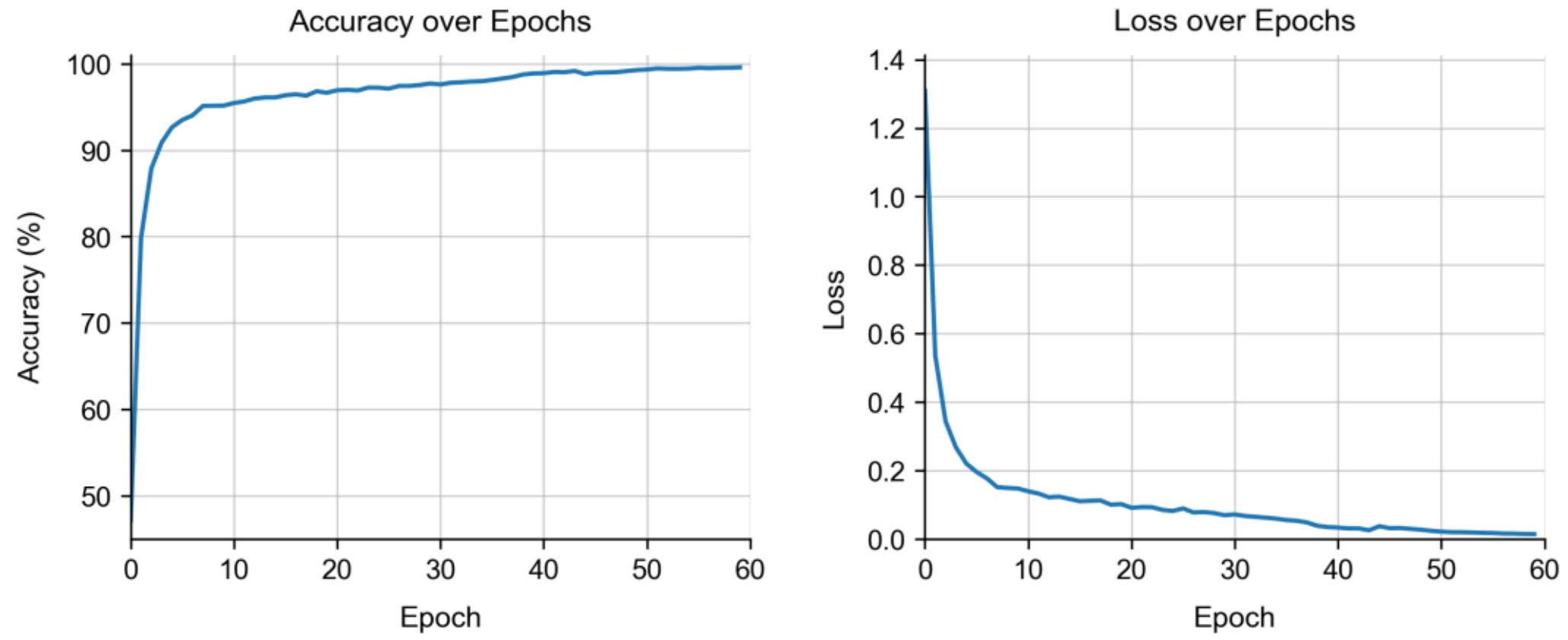

**Fig. 7** Loss and accuracy during the training process

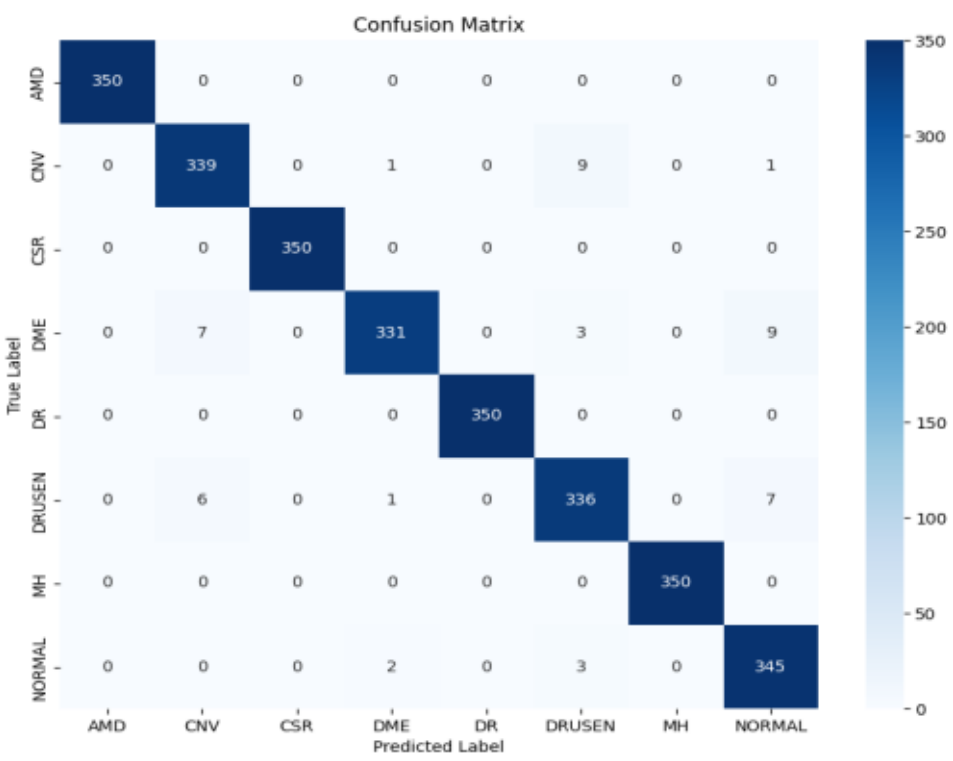

**Fig. 8** Confusion matrix of RetiWave-Mamba on the OCT-C8 test set for a representative run, with an accuracy of 98.25%

The backbone-level complexity results in **Tab. 1** also provide a practical basis for using ResNet-50 in the low-frequency stream. Under the tested setting, ResNet-50 has a lower measured inference time than DenseNet121, EfficientNet-B3, and Swin-Tiny, although DenseNet121 and EfficientNet-B3 require fewer parameters and FLOPs. Together with its staged residual structure, which facilitates the integration of MCLM at successive feature levels, this inference efficiency supports our choice of ResNet-50 as the low-frequency backbone.

To further examine how the Mamba modules scale with input resolution, we profiled RetiWave-Mamba at three input sizes, as summarized in **Tab. 2**. In each FAMP module, the selective scan processes the flattened feature sequence with linear computational complexity in the sequence length [39]. Accordingly, the number of parameters remains constant at 65.92 M, while the estimated FLOPs increase from 7.42 G to 15.04 G and 29.48 G, respectively. When the input side length increases from 224 to 448, the number of pixels increases fourfold, while the FLOPs increase by approximately 3.97 times. These results show that, within the evaluated range, the computational cost of RetiWave-Mamba grows approximately linearly with the number of input pixels rather than quadratically with the sequence length as in self-attention.

**Tab. 2** Computational scaling of RetiWave-Mamba with increasing input size

| Input size | Params (M) | FLOPs (G) | Relative pixels | Relative FLOPs |
|---|---|---|---|---|
| 224 × 224 | 65.92 | 7.42 | 1.00× | 1.00× |
| 320 × 320 | 65.92 | 15.04 | 2.04× | 2.03× |
| 448 × 448 | 65.92 | 29.48 | 4.00× | 3.97× |

### 4.5 Ablation Study

To evaluate the contribution of each module in the proposed model to the overall performance, we conducted systematic ablation experiments on the OCT-C8 dataset, the results of which are shown in **Tab. 3**. The accuracy was 97.59 ± 0.14% when using ResNet-50 alone, which we adopted as our baseline. We first introduced the Discrete Wavelet Transform (DWT) to construct a dual-stream architecture, decomposing the input into low- and high-frequency components. This modification led to an accuracy of 97.71 ± 0.10%, marking a 0.12% improvement over the baseline, suggesting that decoupling structural and textural information facilitates more effective feature extraction. Subsequently, we integrated the Multi-scale Contextual Localization Module (MCLM) into the low-frequency branch. By expanding the receptive field and enhancing lesion localization, this module increased the accuracy to 98.02 ± 0.14%. Finally, we incorporated the complete High-Frequency Branch, comprising the Attention-Guided High-Resolution Network (AG-HRNet) and the Feature-Adaptive Mamba Projector (FAMP). This integration elevated the model's performance to a state-of-the-art level of 98.38 ± 0.12%. The further improvement demonstrates that the high-frequency stream effectively captures fine-grained edge details and long-range textural dependencies that are overlooked by the low-frequency branch. The synergistic operation of all proposed modules ensures that these subtle features are utilized robustly for diagnosis, confirming the necessity of a comprehensive dual-stream design.

**Tab. 3** Ablation experiment results on the OCT-C8 dataset (%)

| Method | Accuracy | Precision | Sensitivity | F1-Score |
|---|---|---|---|---|
| ResNet50[9] | 97.59 ± 0.14 | 97.60 ± 0.13 | 97.59 ± 0.14 | 97.59 ± 0.14 |
| ResNet50+DWT | 97.71 ± 0.10 | 97.72 ± 0.10 | 97.71 ± 0.10 | 97.71 ± 0.10 |
| ResNet50+DWT+MCLM | 98.02 ± 0.14 | 98.03 ± 0.13 | 98.02 ± 0.14 | 98.02 ± 0.14 |
| **RetiWave-Mamba (ours)** | **98.38 ± 0.12** | **98.39 ± 0.11** | **98.38 ± 0.12** | **98.38 ± 0.12** |

To investigate the effect of different context aggregation designs within MCLM, we compared MSW_SA + ASPP [56], MSW_SA + RFB [57], and the proposed MSW_SA + MDF combination. Only the context aggregation module was replaced, while MSW_SA and the remaining network components were kept unchanged. As shown in **Tab. 4**, MSW_SA + MDF achieved the highest mean accuracy and F1 score, both at 98.38 ± 0.12%. MSW_SA + ASPP and MSW_SA + RFB obtained mean accuracies of 98.17 ± 0.16% and 98.35 ± 0.12%, respectively. Compared with MSW_SA + ASPP, the proposed combination achieved higher mean classification performance with fewer parameters and FLOPs. MSW_SA + RFB had the lowest parameter count and FLOPs, while MSW_SA + MDF achieved slightly higher mean classification performance. These results support the effectiveness of combining MDF with MSW_SA for contextual feature extraction within MCLM.

**Tab. 4** Comparison of different context aggregation modules on the OCT-C8 dataset

| Method | Accuracy | Precision | Sensitivity | F1-Score | Params (M) | FLOPs (G) |
|---|---|---|---|---|---|---|
| MSW_SA + ASPP[56] | 98.17 ± 0.16 | 98.17 ± 0.16 | 98.17 ± 0.16 | 98.16 ± 0.16 | 142.63 | 27.14 |
| MSW_SA + RFB[57] | 98.35 ± 0.12 | 98.35 ± 0.12 | 98.35 ± 0.12 | 98.34 ± 0.12 | **20.37** | **3.99** |
| **MSW_SA + MDF (ours)** | **98.38 ± 0.12** | **98.39 ± 0.11** | **98.38 ± 0.12** | **98.38 ± 0.12** | 33.57 | 6.58 |

The MCLM incorporates the MSW_SA unit to achieve precise lesion localization within the expanded receptive field. To verify its superiority, we benchmarked MSW_SA against two widely adopted attention mechanisms: the Squeeze-and-Excitation (SE) block [25], representing channel attention, and the Convolutional Block Attention Module (CBAM) [26], representing hybrid channel-spatial attention.

**Tab. 5** Comparison of different attention mechanisms on the OCT-C8 dataset (%)

| Method | Accuracy | Precision | Sensitivity | F1-Score |
|---|---|---|---|---|
| MDF + CBAM[26] | 97.99 ± 0.13 | 98.00 ± 0.13 | 97.99 ± 0.13 | 97.99 ± 0.13 |
| MDF + SE[25] | 98.14 ± 0.09 | 98.15 ± 0.09 | 98.14 ± 0.09 | 98.14 ± 0.09 |
| **MDF + MSW_SA (ours)** | **98.38 ± 0.12** | **98.39 ± 0.11** | **98.38 ± 0.12** | **98.38 ± 0.12** |

As presented in **Tab. 5**, MDF + MSW_SA achieved the best performance with an accuracy of 98.38 ± 0.12%. In contrast, MDF + SE and MDF + CBAM obtained mean accuracies of 98.14 ± 0.09% and 97.99 ± 0.13%, respectively. The performance gap can be attributed to the specific challenges of OCT imaging. The SE block focuses solely on channel interdependencies, ignoring the spatial distribution of lesions, which limits its localization capability. Although CBAM incorporates a spatial attention module, it relies on standard convolutions that treat all spatial pixels equally. Consequently, it is susceptible to the severe speckle noise inherent in OCT images, often leading to attention drifting toward background artifacts. Our MSW_SA overcomes these limitations by integrating wavelet transforms into the attention generation process. This allows the module to inherently filter out high-frequency noise while capturing multi-scale structural information, ensuring the model focuses precisely on pathological regions.

To evaluate the contributions of AG-HRNet and FAMP to the high-frequency branch, we compared AG-HRNet + Standard Mamba, Standard HRNet + FAMP, and the proposed AG-HRNet + FAMP variant. As shown in **Tab. 6**, with FAMP fixed, AG-HRNet + FAMP achieved an accuracy of 98.38 ± 0.12%, which was comparable to the 98.38 ± 0.11% obtained by Standard HRNet + FAMP. With AG-HRNet fixed, replacing Standard Mamba with FAMP increased the mean accuracy and F1 score from 98.33 ± 0.10% to 98.38 ± 0.12%. Overall, the three configurations achieved similar results, showing that the complete high-frequency configuration maintains comparable overall classification performance to its standard counterparts.

**Tab. 6** Performance comparison of high-frequency branch variants on the OCT-C8 dataset (%)

| Method | Accuracy | Precision | Sensitivity | F1-Score |
|---|---|---|---|---|
| AG-HRNet + Standard Mamba | 98.33 ± 0.10 | 98.35 ± 0.10 | 98.33 ± 0.10 | 98.33 ± 0.10 |
| Standard HRNet + FAMP | 98.38 ± 0.11 | 98.39 ± 0.11 | 98.38 ± 0.11 | 98.37 ± 0.11 |
| **AG-HRNet + FAMP (ours)** | **98.38 ± 0.12** | **98.39 ± 0.11** | **98.38 ± 0.12** | **98.38 ± 0.12** |

**Tab. 7** Performance comparison of high-frequency branch variants on the structurally similar CNV and DRUSEN classes in OCT-C8 (%)

| Method | Accuracy | Precision | Sensitivity | F1-Score |
|---|---|---|---|---|
| AG-HRNet + Standard Mamba | 96.00 ± 0.37 | **96.63 ± 0.34** | 96.00 ± 0.37 | 96.31 ± 0.33 |
| Standard HRNet + FAMP | 96.12 ± 0.41 | 96.54 ± 0.28 | 96.12 ± 0.41 | **96.33 ± 0.33** |
| **AG-HRNet + FAMP (ours)** | **96.21 ± 0.37** | 96.43 ± 0.28 | **96.21 ± 0.37** | 96.32 ± 0.22 |

Since overall metrics may not fully capture performance on structurally similar classes, we further conducted a targeted quantitative evaluation of the three variants on the CNV and DRUSEN subset, as shown in **Tab. 7**. In this evaluation, CNV and DRUSEN retained their original labels rather than being merged into a single category. Accuracy was calculated over all CNV and DRUSEN samples, while precision, sensitivity, and F1 were macro-averaged across the two classes. AG-HRNet + FAMP achieved the highest accuracy and sensitivity, both at 96.21 ± 0.37%, while AG-HRNet + Standard Mamba obtained the highest precision of 96.63 ± 0.34%. Standard HRNet + FAMP achieved the highest F1 score of 96.33 ± 0.33%, which was close to the 96.32 ± 0.22% obtained by the proposed AG-HRNet + FAMP variant. Moreover, this variant had the lowest standard deviation in F1 score, suggesting more consistent performance across repeated runs. Overall, AG-HRNet + FAMP achieved the highest accuracy and sensitivity while maintaining a competitive F1 score, demonstrating balanced and stable performance on the structurally similar CNV and DRUSEN classes.

### 4.6 Robustness Analysis

OCT image quality varies significantly because the acquisition process is inevitably affected by external factors, such as imaging equipment performance and ambient light interference. Furthermore, due to interference signals caused by the back-scattered light of biological tissues, original OCT images often contain severe speckle noise. These issues weaken the edges and fine details of the retinal layers, posing major challenges for subsequent feature extraction and diagnosis. To evaluate and improve model robustness under controlled synthetic speckle-noise conditions, we implemented a three-stage progressive noise-injection training protocol. This protocol was conducted separately from the standard training used to obtain the results reported in **Tab. 1**. We used a clipped multiplicative noise model to keep pixel values within the valid range:

$$I_{aug} = \mathrm{Clamp}\big(I_{clean} \cdot \big(1 + \mathcal{N}(0, \sigma^2)\big), 0,1\big) \tag{23}$$

By keeping pixel values in a valid range, this formula prevents the model from learning unrealistic features. Here, $p$ denotes the probability that noise is applied to a training image, and, when noise is applied, $\sigma$ is independently sampled from a uniform distribution over the interval specified for the current phase. Furthermore, we dynamically regulated the training process across three distinct phases to progressively build the model's tolerance to noise. In Phase 1 (Initial Stabilization, Epochs 0–20), we applied low noise ($p = 0.5$, $\sigma \in [0.0,0.1]$) to help the network learn basic structural features stably. Subsequently, in Phase 2 (Progressive Adaptation, Epochs 20–40), we increased the difficulty ($p = 0.8$, $\sigma \in [0.05,0.15]$), preventing reliance on clear edges and forcing the model to adapt to textural corruption. Finally, in Phase 3 (High-Intensity Optimization, Epochs 40–60), we implemented a fully noisy training phase ($p = 1.0$, $\sigma \in [0.1,0.2]$) using only degraded data. This rigorous condition compelled the network to extract discriminative features even from heavily corrupted inputs.

To verify the effectiveness of our RetiWave-Mamba model in noisy environments, we added a multiplicative scattering noise model [17] to the test dataset. The noise generation process and the Peak Signal-to-Noise Ratio (PSNR) measurement are expressed by the following equations:

$$F(x,y) = g(x,y) + g(x,y) \times u(x,y) \tag{24}$$

$$PSNR = 20 \times \log_{10}\left(\frac{MAX}{\sqrt{MSE}}\right) \tag{25}$$

where $g(x,y)$ denotes the clean image, $u(x,y)$ is Gaussian noise with variance $\sigma^2$ dependent on gray values, and $F(x,y)$ is the noisy image. In Eq. (25), $MAX$ is the maximum pixel value and $MSE$ is the mean square error. After completing the three-stage progressive noise-injection training protocol, the parameters of each trained model were kept unchanged. For each run, the same trained model was evaluated on the OCT-C8 test set under four conditions: the original noiseless condition and three noise-corrupted conditions corresponding to PSNR values of 32.19, 28.69, and 26.22 dB. No retraining, fine-tuning, or test-time adaptation was performed between the four test conditions. To ensure a fair comparison, all models included in the robustness analysis were trained using the same three-stage progressive noise-injection schedule and evaluated using identical test images, noise-generation procedures, and noise levels. All results are reported as the mean ± standard deviation over six independent runs.

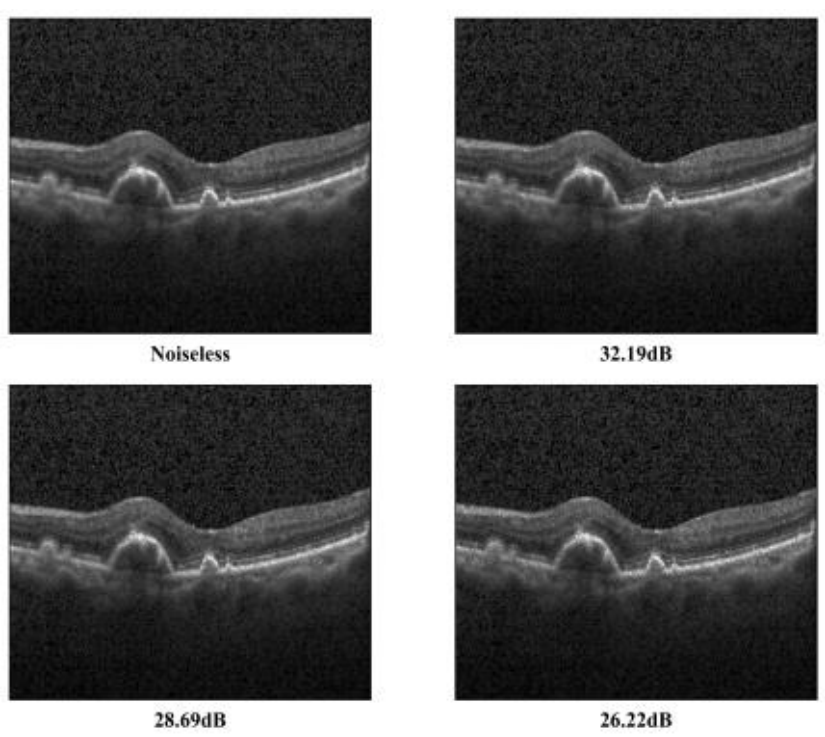

**Fig. 9** Visualization of OCT-C8 test samples before and after adding speckle noise at varying intensities

**Fig. 9** presents representative OCT-C8 test samples before and after the addition of speckle noise at varying intensities. It can be observed that as the noise intensity increases, the boundary definitions between retinal layers become increasingly blurred, and fine pathological details are obscured. These observations show that the synthetic perturbation produces controlled degradation of retinal structures as the noise level increases.

As shown in **Tab. 8**, RetiWave-Mamba achieved the highest accuracy under the noiseless condition and at PSNR levels of 32.19 dB and 28.69 dB, with accuracies of 98.24 ± 0.13%, 98.25 ± 0.10%, and 97.99 ± 0.15%, respectively. At the highest noise level of 26.22 dB, its accuracy was 97.41 ± 0.20%, while WaveNet-SF achieved the highest accuracy of 97.77 ± 0.18%. Swin-Tiny decreased from 97.55 ± 0.08% under the noiseless condition to 96.87 ± 0.21% at 26.22 dB. These results indicate consistent robustness to the evaluated synthetic speckle-noise levels, although RetiWave-Mamba did not achieve the highest accuracy under the strongest noise condition.

**Tab. 8** Classification accuracy of different models on the clean and noise-corrupted OCT-C8 test sets after three-stage progressive noise-injection training (%)

| Method \ Noise intensity | Noiseless | 32.19dB | 28.69dB | 26.22dB |
|---|---|---|---|---|
| VGG16[10] | 97.82 ± 0.08 | 97.72 ± 0.06 | 97.65 ± 0.12 | 97.45 ± 0.14 |
| GoogLeNet[49] | 97.86 ± 0.17 | 97.76 ± 0.17 | 97.70 ± 0.11 | 97.58 ± 0.11 |
| ResNet50[9] | 97.77 ± 0.16 | 97.66 ± 0.17 | 97.75 ± 0.18 | 97.41 ± 0.30 |
| InceptionV3[51] | 97.80 ± 0.12 | 97.81 ± 0.20 | 97.68 ± 0.17 | 97.54 ± 0.17 |
| DenseNet121[50] | 97.39 ± 0.28 | 97.45 ± 0.15 | 97.38 ± 0.18 | 97.35 ± 0.18 |
| EfficientNet-B3[52] | 96.36 ± 0.15 | 96.37 ± 0.17 | 96.49 ± 0.21 | 96.20 ± 0.27 |
| Swin-Tiny[54] | 97.55 ± 0.08 | 97.42 ± 0.23 | 97.40 ± 0.16 | 96.87 ± 0.21 |
| ConvNeXt V2[53] | 95.66 ± 0.33 | 95.57 ± 0.29 | 95.50 ± 0.34 | 95.43 ± 0.33 |
| MSLI-Net[15] | 97.32 ± 0.12 | 97.35 ± 0.07 | 97.18 ± 0.19 | 97.15 ± 0.18 |
| WaveNet-SF[14] | 97.74 ± 0.15 | 97.79 ± 0.12 | 97.72 ± 0.08 | **97.77 ± 0.18** |
| **RetiWave-Mamba (ours)** | **98.24 ± 0.13** | **98.25 ± 0.10** | **97.99 ± 0.15** | 97.41 ± 0.20 |

### 4.7 Evaluation of model generalization ability

To evaluate performance beyond OCT-C8, we directly applied the OCT-C8-trained models to the OCT2017 test set [58] without additional training or fine-tuning. This test set contains 1,000 images, with 250 images in each of four categories: CNV, DME, DRUSEN, and Normal. To align the label spaces, predictions were restricted to the four corresponding outputs of the original eight-class classifiers. The same evaluation procedure was applied to all compared models, and results are reported as mean ± standard deviation across the six OCT-C8-trained checkpoints.

**Tab. 9** Performance comparison on the OCT2017 dataset (%)

| Method | Accuracy | Precision | Sensitivity | F1-Score |
|---|---|---|---|---|
| VGG16[10] | 99.47 ± 0.21 | 99.47 ± 0.20 | 99.47 ± 0.21 | 99.47 ± 0.21 |
| GoogLeNet[49] | **99.52 ± 0.26** | **99.52 ± 0.26** | **99.52 ± 0.26** | **99.52 ± 0.26** |
| ResNet50[9] | 99.48 ± 0.21 | 99.48 ± 0.21 | 99.48 ± 0.21 | 99.48 ± 0.21 |
| DenseNet121[50] | 99.25 ± 0.15 | 99.25 ± 0.15 | 99.25 ± 0.15 | 99.25 ± 0.15 |
| EfficientNet-B3[52] | 99.38 ± 0.10 | 99.39 ± 0.10 | 99.38 ± 0.10 | 99.38 ± 0.10 |
| Swin-Tiny[54] | 99.20 ± 0.11 | 99.20 ± 0.11 | 99.20 ± 0.11 | 99.20 ± 0.11 |
| ConvNeXt V2[53] | 99.32 ± 0.12 | 99.32 ± 0.12 | 99.32 ± 0.12 | 99.32 ± 0.12 |
| WaveNet-SF[14] | 99.47 ± 0.14 | 99.47 ± 0.14 | 99.47 ± 0.14 | 99.47 ± 0.14 |
| **RetiWave-Mamba (ours)** | 99.43 ± 0.14 | 99.44 ± 0.13 | 99.43 ± 0.14 | 99.43 ± 0.14 |

As shown in **Tab. 9**, RetiWave-Mamba achieved an accuracy of 99.43 ± 0.14%, a precision of 99.44 ± 0.13%, and sensitivity and F1 scores of 99.43 ± 0.14% on the OCT2017 test set. Its mean accuracy was higher than those of DenseNet121, EfficientNet-B3, Swin-Tiny, and ConvNeXt V2, but slightly lower than those of GoogLeNet, ResNet50, VGG16, and WaveNet-SF. These results support the cross-dataset generalization ability of RetiWave-Mamba, which maintains high classification accuracy on OCT2017 without target-dataset fine-tuning.

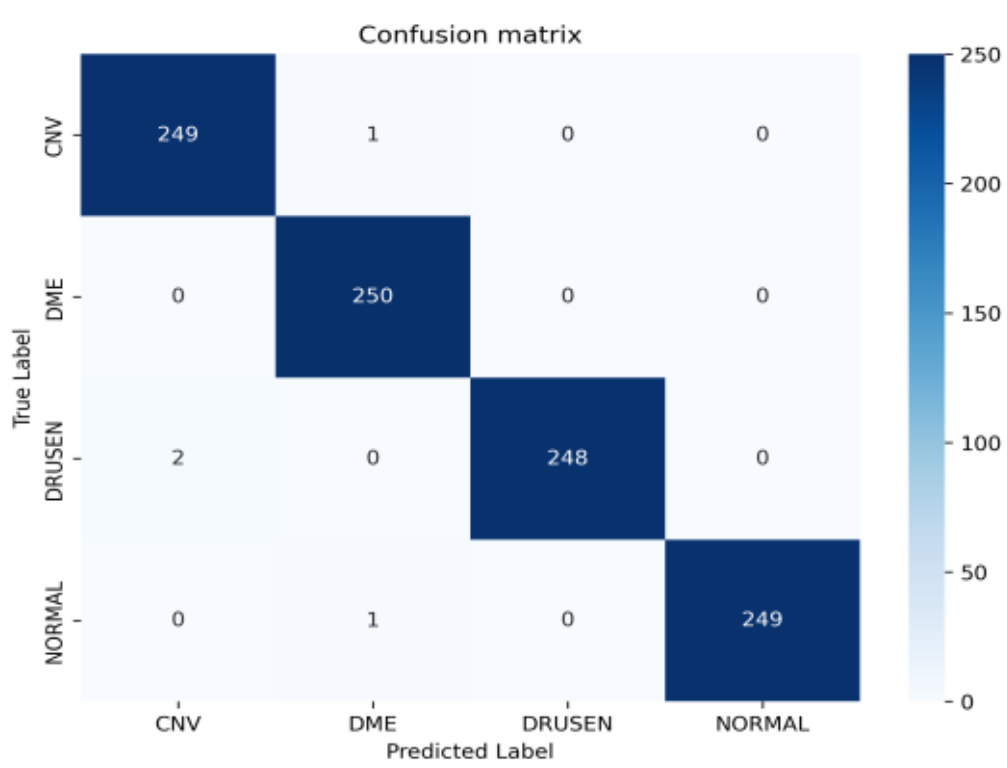

**Fig. 10** Confusion matrix of RetiWave-Mamba on the OCT2017 test set for a representative run

To further visualize the classification performance on this external dataset, **Fig. 10** presents the confusion matrix. It can be observed that the model achieves near-perfect classification across all four categories, with minimal misclassifications. Specifically, the model effectively distinguishes between pathological cases and normal eyes, as well as between different disease types, despite the domain gap. These results indicate that our dual-stream design effectively learns invariant pathological features—structural anomalies via the MCLM and fine-grained textures via the FAMP—rather than overfitting to the specific artifacts of the source dataset. This cross-dataset generalization suggests the potential of RetiWave-Mamba for computer-aided retinal disease classification under different data distributions.

### 4.8 Prediction calibration analysis

To assess the reliability of the predicted probabilities, we evaluated the calibration performance of RetiWave-Mamba and ten comparison methods on the OCT-C8 test set. Class probabilities were obtained by applying softmax to the original model outputs, without post-hoc calibration. Mean confidence, expected calibration error (ECE), negative log-likelihood (NLL), and the multiclass Brier score are reported as mean ± standard deviation over six runs. ECE was calculated using 15 equal-width confidence bins. The Brier score was computed by summing the squared probability errors across classes and averaging over test samples.

As shown in **Tab. 10**, RetiWave-Mamba achieved the lowest mean ECE, NLL, and Brier score among the evaluated methods, at 1.18 ± 0.11%, 0.0735 ± 0.0047, and 0.0283 ± 0.0019, respectively. Its ECE was lower than those of EfficientNet-B3 (1.30 ± 0.21%), Swin-Tiny (1.32 ± 0.10%), and WaveNet-SF (1.54 ± 0.09%). It also obtained lower NLL and Brier scores than these methods. These results support the calibration performance of RetiWave-Mamba on the evaluated test set. Its mean confidence of 99.37 ± 0.09%, however, remained above its accuracy of 98.38 ± 0.12%, indicating some residual overconfidence.

**Tab. 10** Comparison of prediction calibration on the OCT-C8 dataset

| Method | Mean confidence (%) | ECE (%) ↓ | NLL ↓ | Brier score ↓ |
|---|---|---|---|---|
| VGG16[10] | **99.38 ± 0.10** | 1.75 ± 0.11 | 0.1439 ± 0.0149 | 0.0392 ± 0.0016 |
| GoogLeNet[49] | 99.30 ± 0.09 | 1.59 ± 0.11 | 0.0996 ± 0.0096 | 0.0363 ± 0.0011 |
| ResNet50[9] | 99.11 ± 0.13 | 1.59 ± 0.18 | 0.1017 ± 0.0089 | 0.0403 ± 0.0023 |
| InceptionV3[51] | 99.58 ± 0.12 | 1.48 ± 0.15 | 0.0975 ± 0.0069 | 0.0322 ± 0.0026 |
| DenseNet121[50] | 99.28 ± 0.10 | 1.45 ± 0.12 | 0.0913 ± 0.0089 | 0.0348 ± 0.0017 |
| EfficientNet-B3[52] | 98.85 ± 0.10 | 1.30 ± 0.21 | 0.0891 ± 0.0053 | 0.0371 ± 0.0014 |
| Swin-Tiny[54] | 99.09 ± 0.09 | 1.32 ± 0.10 | 0.0838 ± 0.0040 | 0.0349 ± 0.0008 |
| ConvNeXt V2[53] | 98.81 ± 0.04 | 1.69 ± 0.20 | 0.1079 ± 0.0090 | 0.0456 ± 0.0034 |
| MSLI-Net[15] | 99.00 ± 0.11 | 1.68 ± 0.20 | 0.1231 ± 0.0092 | 0.0432 ± 0.0030 |
| WaveNet-SF[14] | 99.22 ± 0.09 | 1.54 ± 0.09 | 0.0983 ± 0.0069 | 0.0380 ± 0.0019 |
| **RetiWave-Mamba (ours)** | 99.37 ± 0.09 | **1.18 ± 0.11** | **0.0735 ± 0.0047** | **0.0283 ± 0.0019** |

# 5. Discussion

## 5.1 The Performance of RetiWave-Mamba

As shown in the experimental results in Section 4.4, we compared our method with a wide range of models, including classic CNNs, modern Transformers, and specialized retinal networks. However, these existing methods often struggle with speckle noise and similarities between different classes. This limit is especially clear when distinguishing between Choroidal Neovascularization (CNV) and DRUSEN, as these lesions look very similar. In contrast, our proposed RetiWave-Mamba framework effectively overcomes these challenges. It achieves a state-of-the-art (SOTA) accuracy of 98.38 ± 0.12% on the OCT-C8 dataset, performing better than previous methods even on these difficult categories.

This superior performance comes from our new dual-stream spatial-frequency design. By using the Discrete Wavelet Transform (DWT), we separate the input image into different frequency parts. This allows us to balance noise suppression and detail preservation. The low-frequency stream uses the Multi-scale Contextual Localization Module (MCLM) to accurately locate the lesion. Meanwhile, the high-frequency stream uses the Attention-Guided High-Resolution Network (AG-HRNet) and the Feature-Adaptive Mamba Projector (FAMP). These components capture the fine edge details and long-range dependencies needed for precise diagnosis.

To provide an intuitive and interpretable evaluation of the model's decision-making process, we utilized the Grad-CAM method [59] to generate heatmaps, as shown in **Fig. 11**. This visualization reveals significant differences in how each model focuses on pathological features under noisy conditions.

In general, for ConvNeXt V2, although it generally locates the lesion area, its activation maps tend to be diffuse and over-smoothed, lacking precise boundary definition. The ResNet50 baseline exhibits a susceptibility to background interference, often drifting its attention towards non-pathological areas or high-contrast artifacts, as clearly seen in AMD (a). In contrast, RetiWave-Mamba consistently generates tightly constrained activation maps that accurately delineate lesion boundaries, thanks to the global context capture of the MCLM module.

It is particularly worth noting the performance on the morphologically similar categories of CNV (b) and DRUSEN (f), which serve as excellent examples of the model's discriminative capability. In these challenging cases, RetiWave-Mamba demonstrates exceptional specificity: it precisely targets the irregular neovascular textures in CNV and the granular depositions in DRUSEN, effectively filtering out the inherent speckle noise. Conversely, the heatmaps of ConvNeXt V2 in these cases are notably blurred and bleed into surrounding healthy tissues, indicating a failure to resolve the fine-grained high-frequency details required to distinguish these RPE elevations. Similarly, ResNet50 struggles significantly here, with its attention drifting towards image noise rather than locking onto the lesion core. This comparison validates that the AG-HRNet and FAMP in our high-frequency branch are crucial for capturing the subtle textural dependencies that standard backbones miss.

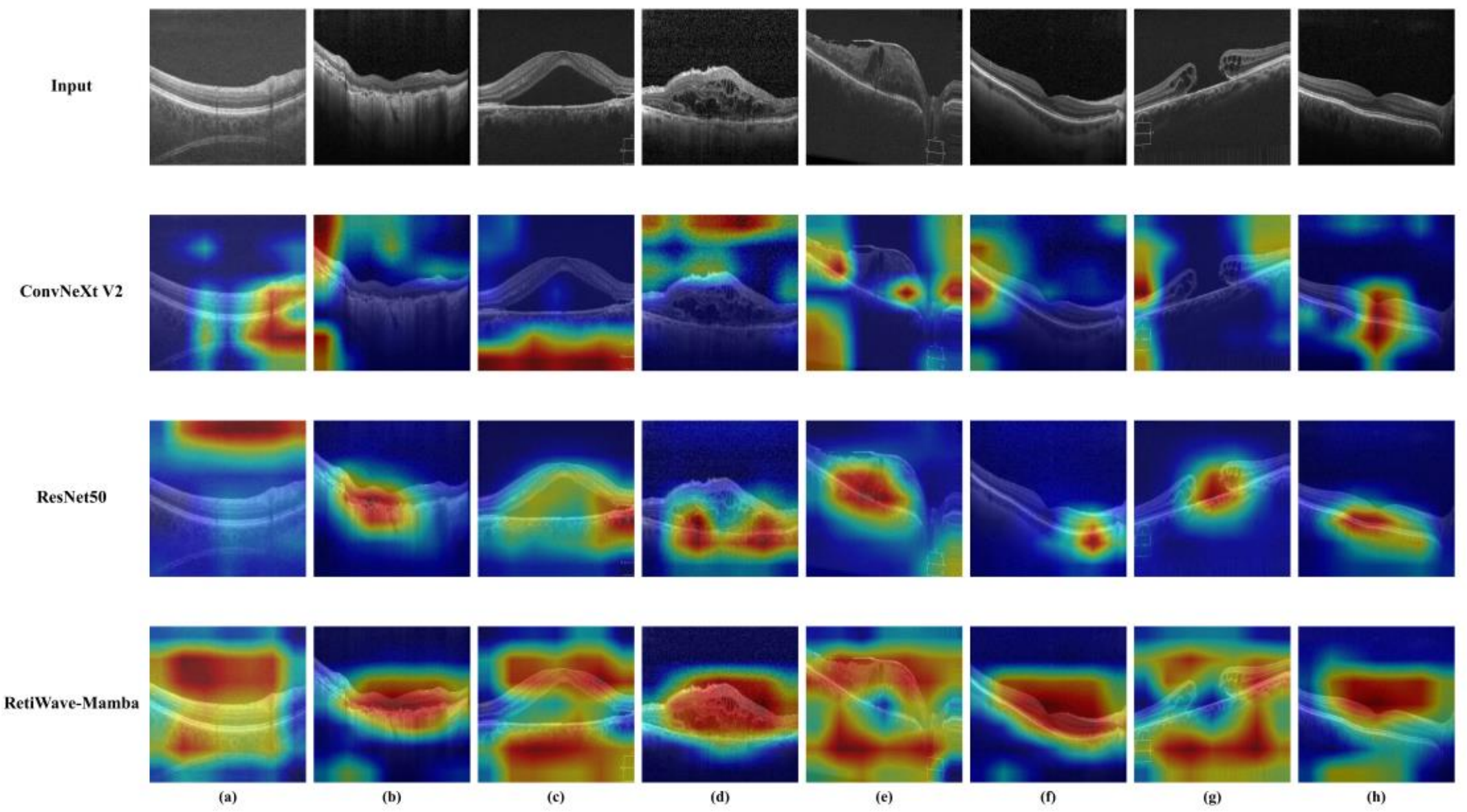

**Fig. 11** Heatmaps of retinal OCT images. (a) AMD (b) CNV (c) CSR (d) DME (e) DR (f) DRUSEN (g) MH (h) NORMAL

### 5.2 Ablation Study

In Section 4.5, we quantitatively verified the contribution of each proposed module through stepwise ablation. To provide a deeper understanding of why our specific designs—namely the Multi-scale Contextual Localization Module (MCLM) and the core high-frequency components (AG-HRNet and FAMP)—outperform traditional approaches, we analyze their internal mechanisms and comparative performance in detail below.

#### *5.2.1 Efficacy of the MCLM Components: MDF and MSW_SA*

To intuitively validate the rationality of the proposed Multi-scale Contextual Localization Module (MCLM) and demonstrate its superiority over other attention mechanisms, we employed LayerCAM [60] to visualize the feature evolution within the low-frequency branch. **Fig. 12** illustrates the intermediate feature maps from Stage 1 to Stage 4, alternating between the raw backbone features and the refined features output by the attention modules. As observed in the horizontal evolution, the raw features from the backbone typically contain significant background noise, such as high-reflective retinal layers and vitreous artifacts. However, after passing through our attention modules, the background clutter is progressively suppressed, and the focus shifts sharply toward the pathological regions. This phenomenon confirms that the MCLM acts as an effective filter, highlighting discriminative lesion features while discarding irrelevant noise layer by layer.

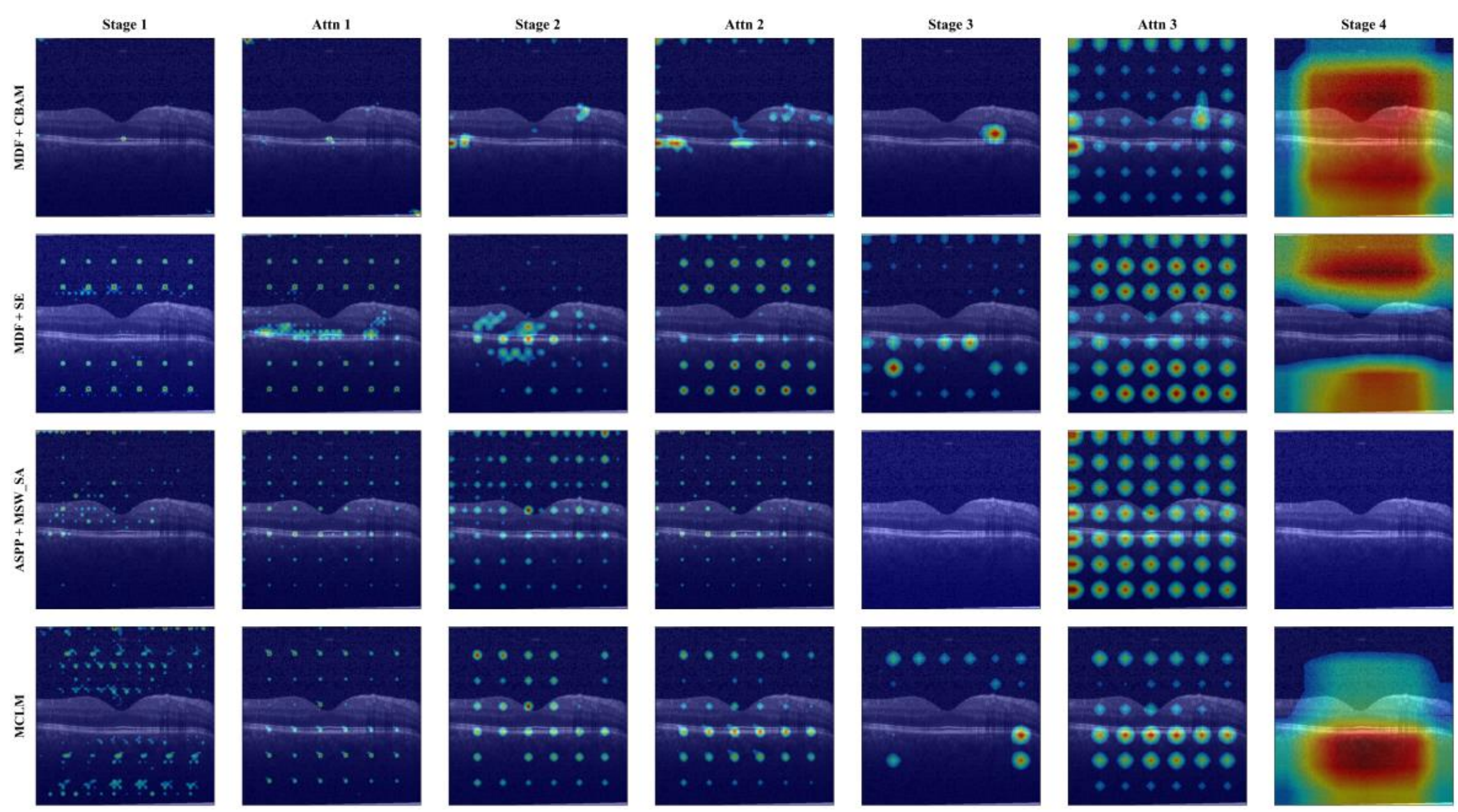

**Fig. 12** LayerCAM visualization comparison of feature evolution in the low-frequency branch

A detailed comparative analysis reveals critical limitations in existing mechanisms when applied to OCT imagery. Regarding the "MDF + SE" variant, visual inspection reveals that the attention maps exhibit a significant deviation from the pathological region. High activation areas are erroneously concentrated in the background regions above and below the main retinal structure corresponding to the vitreous humor and sclera, indicating that channel attention alone fails to spatially filter out background noise. Similarly, for the "MDF + CBAM" configuration, although the focus on the main structure is improved, it remains susceptible to the inherent speckle noise in OCT images. This manifests as false activations in the upper non-pathological regions like the choroid, suggesting that standard spatial convolutions struggle to distinguish between pathological textures and high-frequency speckle noise. Furthermore, the "MSW_SA + ASPP" approach exposes severe structural limitations in deep feature extraction: the Attn 3 map presents a distinct lattice-like activation pattern, while the Stage 4 map suffers from complete signal loss, appearing entirely dark. This signal vanishing is attributed to the mismatch between the large dilation rates of ASPP and the reduced spatial resolution of deep feature maps.

In contrast, our proposed MCLM effectively overcomes these issues through the synergistic combination of MDF and MSW_SA. While displaying a lattice-like activation pattern in intermediate layers, the additive integration of multi-scale contextual modeling (MDF) and wavelet-based frequency filtering (MSW_SA) allows the model to maintain significantly more distinct and coherent highlighting of the retinal structure. This combined mechanism ensures that pathological features remain dominant over texture responses. Most importantly, unlike the dilation-based "MSW_SA + ASPP" baseline, this dual-module design successfully preserves strong semantic signals in deep layers (Stage 4), effectively suppressing background noise and avoiding false choroidal activations. The visualization confirms that the strategic combination of multi-scale fusion and wavelet-based attention achieves the most precise and robust lesion localization.

#### *5.2.2 Efficacy of the High-Frequency Branch Components*

The High-Frequency Branch is designed to preserve fine-grained texture and boundary information,

which is important for distinguishing retinal diseases with similar structural patterns. To evaluate the contributions of AG-HRNet and FAMP, we compared the complete AG-HRNet + FAMP configuration with Standard HRNet + FAMP and AG-HRNet + Standard Mamba, respectively.

As shown in **Tab. 6**, AG-HRNet + FAMP and Standard HRNet + FAMP achieved accuracies of 98.38 ± 0.12% and 98.38 ± 0.11%, respectively, while the proposed AG-HRNet + FAMP variant obtained a slightly higher mean F1 score of 98.38 ± 0.12%, compared with 98.37 ± 0.11% for Standard HRNet + FAMP. These results show that AG-HRNet maintains competitive overall classification performance while using learned gates to control the exchange of multi-resolution high-frequency information. With AG-HRNet fixed, replacing Standard Mamba with FAMP increased the mean accuracy and F1 score from 98.33 ± 0.10% to 98.38 ± 0.12%. Similar improvements were observed in precision and sensitivity, suggesting that FAMP can adaptively reorganize and reweight high-frequency features without reducing overall performance.

To further examine performance on structurally similar classes, **Tab. 7** presents a targeted comparison on the CNV and DRUSEN subset. AG-HRNet + FAMP achieved the highest accuracy and sensitivity, both at 96.21 ± 0.37%, while maintaining a competitive F1 score of 96.32 ± 0.22%. In addition, the proposed AG-HRNet + FAMP variant had the lowest standard deviation in F1 score, indicating more consistent performance across repeated runs. These results suggest that AG-HRNet + FAMP provides a balanced and stable representation for the structurally similar CNV and DRUSEN classes. Specifically, AG-HRNet preserves multi-resolution boundary information, while FAMP adaptively projects and reweights the resulting high-frequency features before fusion. Their complementary roles support the recognition of subtle differences between similar retinal disease classes.

### 5.3 Robustness of RetiWave-Mamba to Noise

OCT images may be affected by device-dependent speckle noise, making robustness to image degradation important for reliable automated analysis. We therefore evaluated RetiWave-Mamba using the progressive noise-injection protocol described in Section 4.6. The protocol creates controlled synthetic speckle-noise levels and does not aim to reproduce every source of variation in clinical imaging.

As shown in **Tab. 8**, RetiWave-Mamba achieved the highest accuracy under the noiseless condition and at PSNR levels of 32.19 dB and 28.69 dB, with mean accuracies of 98.24 ± 0.13%, 98.25 ± 0.10%, and 97.99 ± 0.15%, respectively. At the highest noise level of 26.22 dB, its mean accuracy was 97.41 ± 0.20%, while WaveNet-SF achieved the highest mean accuracy of 97.77 ± 0.18%. Overall, RetiWave-Mamba maintained a mean accuracy above 97.4% across all tested conditions, with a maximum decrease of 0.83 percentage points from the noiseless condition to the highest noise level.

All methods in this comparison were trained using the same progressive noise-injection schedule and evaluated on the same OCT-C8 test set with identical noise-generation procedures and noise levels. Across the compared methods, the effect of increasing noise varied. For example, Swin-Tiny showed an accuracy decrease of 0.68 percentage points, from 97.55 ± 0.08% under the noiseless condition to 96.87 ± 0.21% at 26.22 dB. In comparison, RetiWave-Mamba achieved the highest mean accuracy under the noiseless condition and at the two intermediate noise levels, and remained competitive at the highest noise level, although it did not achieve the highest accuracy under this strongest noise condition. These results indicate consistent robustness to the evaluated synthetic speckle-noise levels.

### 5.4 Limitations of the Evaluation Protocol

Merging the original validation set into the training set leaves no independent validation set for model selection. Although repeated runs and the OCT2017 evaluation provide evidence of performance variability and cross-dataset performance, they do not replace validation-based model selection. Future evaluations should retain a separate validation set or adopt cross-validation to provide a more rigorous basis for model selection.

## 6. Conclusion

In this study, we proposed RetiWave-Mamba, a novel dual-stream framework that synergizes spatial-frequency domain learning with State Space Models to address the critical challenges of speckle noise and lesion scale variability in OCT image analysis. Specifically, by utilizing DWT for spectral decoupling, the framework coordinates the MCLM in the low-frequency stream for precise lesion localization, while integrating AG-HRNet and FAMP in the high-frequency stream to suppress noise artifacts and capture fine-grained textures. Comprehensive ablation studies further confirmed the validity and contribution of each proposed module. Our model achieves a state-of-the-art accuracy of 98.38 ± 0.12% on the OCT-C8 dataset and maintains high classification accuracy under the evaluated synthetic speckle-noise conditions. These results support the potential of RetiWave-Mamba for computer-aided retinal disease classification. Validation using naturally degraded, multi-device, and multi-center OCT data remains necessary before clinical deployment.

### Declaration of competing interest

The authors declare that there are no conflicts of interest regarding the publication of this paper.

### Acknowledgements

This work was supported in part by the National Natural Science Foundation of China (62466033), and in part by the Jiangxi Provincial Natural Science Foundation (20242BAB20070).

## References

1. Meng, Y., et al., *Global, Regional, and National Burden of Blindness due to Diabetic Retinopathy, 1990–2021.* Ophthalmology and Therapy, 2025. **14**(10): p. 2599-2615.
2. Jeong, Y.D., et al., *Global burden of vision impairment due to age-related macular degeneration, 1990–2021, with forecasts to 2050: a systematic analysis for the Global Burden of Disease Study 2021.* The Lancet Global Health, 2025. **13**(7): p. e1175-e1190.
3. Rusciano, D. and S. Marsili, *Editorial to the Special Issue "Retinopathies: A Challenge for Early Diagnosis, Innovative Treatments, and Reliable Follow-Up"*. 2025, MDPI. p. 662.
4. Sorrentino, F.S., et al., *Novel approaches for early detection of retinal diseases using artificial intelligence.* Journal of Personalized Medicine, 2024. **14**(7): p. 690.
5. Huang, D., et al., *Optical coherence tomography.* science, 1991. **254**(5035): p. 1178-1181.
6. Kermany, D.S., et al., *Identifying medical diagnoses and treatable diseases by image-based deep learning.* cell, 2018. **172**(5): p. 1122-1131. e9.
7. Li, T., et al., *Applications of deep learning in fundus images: A review.* Medical Image Analysis, 2021. **69**: p. 101971.
8. Tsuji, T., et al., *Classification of optical coherence tomography images using a capsule network.* BMC ophthalmology, 2020. **20**(1): p. 114.
9. He, K., et al. *Deep residual learning for image recognition*. in *Proceedings of the IEEE conference on computer vision and pattern recognition (CVPR)*. 2016.
10. Simonyan, K. and A. Zisserman, *Very deep convolutional networks for large-scale image recognition.* arXiv preprint

arXiv:1409.1556, 2014.

11. Huang, L., et al., *Automatic classification of retinal optical coherence tomography images with layer guided convolutional neural network.* IEEE Signal Processing Letters, 2019. **26**(7): p. 1026-1030.
12. Peng, J., et al., *Multi-scale-denoising residual convolutional network for retinal disease classification using OCT.* Sensors, 2023. **24**(1): p. 150.
13. Fang, L., et al., *Attention to lesion: Lesion-aware convolutional neural network for retinal optical coherence tomography image classification.* IEEE transactions on medical imaging, 2019. **38**(8): p. 1959-1970.
14. Cheng, J., et al., *WaveNet-SF: A Hybrid Network for Retinal Disease Detection Based on Wavelet Transform in the Spatial-Frequency Domain learning.* Neural Networks, 2025: p. 108189.
15. Qi, Z., et al., *MSLI-Net: retinal disease detection network based on multi-segment localization and multi-scale interaction.* Frontiers in Cell and Developmental Biology, 2025. **13**: p. 1608325.
16. Karthik, K. and M. Mahadevappa, *Convolution neural networks for optical coherence tomography (OCT) image classification.* Biomedical Signal Processing and Control, 2023. **79**: p. 104176.
17. Schmitt, J.M., S. Xiang, and K.M. Yung, *Speckle in optical coherence tomography.* Journal of biomedical optics, 1999. **4**(1): p. 95-105.
18. Sotoudeh-Paima, S., et al., *Multi-scale convolutional neural network for automated AMD classification using retinal OCT images.* Computers in biology and medicine, 2022. **144**: p. 105368.
19. Ma, Z., et al., *HCTNet: a hybrid ConvNet-transformer network for retinal optical coherence tomography image classification.* Biosensors, 2022. **12**(7): p. 542.
20. Laouarem, A., et al., *Htc-retina: a hybrid retinal diseases classification model using transformer-convolutional neural network from optical coherence tomography images.* Computers in Biology and Medicine, 2024. **178**: p. 108726.
21. Khalil, I., et al., *OCTNet: A modified multi-scale attention feature fusion network with InceptionV3 for retinal OCT image classification.* Mathematics, 2024. **12**(19): p. 3003.
22. Hemalakshmi, G., et al., *Automated retinal disease classification using hybrid transformer model (SViT) using optical coherence tomography images.* Neural Computing and Applications, 2024. **36**(16): p. 9171-9188.
23. Yu, F. and V. Koltun, *Multi-scale context aggregation by dilated convolutions.* arXiv preprint arXiv:1511.07122, 2015.
24. Feng, S., et al., *CPFNet: Context pyramid fusion network for medical image segmentation.* IEEE transactions on medical imaging, 2020. **39**(10): p. 3008-3018.
25. Hu, J., L. Shen, and G. Sun. *Squeeze-and-excitation networks*. in *Proceedings of the IEEE conference on computer vision and pattern recognition*. 2018.
26. Woo, S., et al. *Cbam: Convolutional block attention module*. in *Proceedings of the European conference on computer vision (ECCV)*. 2018.
27. Park, J., et al., *Bam: Bottleneck attention module.* arXiv preprint arXiv:1807.06514, 2018.
28. Hou, Q., D. Zhou, and J. Feng. *Coordinate attention for efficient mobile network design*. in *Proceedings of the IEEE/CVF conference on computer vision and pattern recognition*. 2021.
29. Schlemper, J., et al., *Attention gated networks: Learning to leverage salient regions in medical images.* Medical image analysis, 2019. **53**: p. 197-207.
30. Zhang, Q., et al., *Beyond being wise after the event: Combining spatial, temporal and spectral information for Himawari-8 early-stage wildfire detection.* International Journal of Applied Earth Observation and Geoinformation, 2023. **124**: p. 103506.
31. Zhang, Q., et al., *10-minute forest early wildfire detection: Fusing multi-type and multi-source information via recursive transformer.* Neurocomputing, 2025. **616**: p. 128963.
32. Zhang, B., et al., *Immediate remote sensing: Dynamic context-adaptive fusion for himawari-8/9 10-minute wildfire detection.* Remote Sensing of Environment, 2026. **344**: p. 115524.

33. Xiao, Y., et al., *TTST: A top-k token selective transformer for remote sensing image super-resolution.* IEEE Transactions on Image Processing, 2024. **33**: p. 738-752.
34. Xiao, Y., et al., *Spiking meets attention: Efficient remote sensing image super-resolution with attention spiking neural networks.* Advances in Neural Information Processing Systems, 2026. **38**: p. 65240-65259.
35. Xu, G., et al., *Haar wavelet downsampling: A simple but effective downsampling module for semantic segmentation.* Pattern recognition, 2023. **143**: p. 109819.
36. Yao, T., et al. *Wave-vit: Unifying wavelet and transformers for visual representation learning*. in *European conference on computer vision*. 2022. Springer.
37. Xiao, Y., et al., *Frequency-assisted mamba for remote sensing image super-resolution.* IEEE Transactions on Multimedia, 2024. **27**: p. 1783-1796.
38. Gu, A., K. Goel, and C. Ré, *Efficiently modeling long sequences with structured state spaces.* arXiv preprint arXiv:2111.00396, 2021.
39. Gu, A. and T. Dao. *Mamba: Linear-time sequence modeling with selective state spaces*. in *First conference on language modeling*. 2024.
40. Zhu, L., et al., *Vision mamba: Efficient visual representation learning with bidirectional state space model.* arXiv preprint arXiv:2401.09417, 2024.
41. Liu, Y., et al., *Vmamba: Visual state space model.* Advances in neural information processing systems, 2024. **37**: p. 103031-103063.
42. Zuo, Q., et al., *Multi-resolution visual Mamba with multi-directional selective mechanism for retinal disease detection.* Frontiers in Cell and Developmental Biology, 2024. **12**: p. 1484880.
43. Jiang, K., et al., *Vdmamba: Vector decomposition in vision mamba for image deraining and beyond.* IEEE Transactions on Multimedia, 2026.
44. Zheng, T., et al., *GraphMamba: Whole slide image classification meets graph-driven selective state space model.* Pattern Recognition, 2025. **167**: p. 111768.
45. Jiang, K., et al., *Ph-mamba: Enhancing mamba with position encoding and harmonized attention for image deraining and beyond.* IEEE Transactions on Image Processing, 2025.
46. Daubechies, I., *Ten lectures on wavelets*. 1992: SIAM.
47. Sun, K., et al. *Deep high-resolution representation learning for human pose estimation*. in *Proceedings of the IEEE/CVF conference on computer vision and pattern recognition*. 2019.
48. Subramanian, M., et al. *Classification of retinal oct images using deep learning*. in *2022 international conference on computer communication and informatics (ICCCI)*. 2022. IEEE.
49. Szegedy, C., et al. *Going deeper with convolutions*. in *Proceedings of the IEEE conference on computer vision and pattern recognition*. 2015.
50. Huang, G., et al. *Densely connected convolutional networks*. in *Proceedings of the IEEE conference on computer vision and pattern recognition*. 2017.
51. Szegedy, C., et al. *Rethinking the inception architecture for computer vision*. in *Proceedings of the IEEE conference on computer vision and pattern recognition*. 2016.
52. Tan, M. and Q. Le. *Efficientnet: Rethinking model scaling for convolutional neural networks*. in *International conference on machine learning*. 2019. PMLR.
53. Woo, S., et al. *Convnext v2: Co-designing and scaling convnets with masked autoencoders*. in *Proceedings of the IEEE/CVF conference on computer vision and pattern recognition*. 2023.
54. Liu, Z., et al. *Swin transformer: Hierarchical vision transformer using shifted windows*. in *Proceedings of the IEEE/CVF international conference on computer vision*. 2021.
55. He, J., et al., *An interpretable transformer network for the retinal disease classification using optical coherence*

*tomography.* Scientific Reports, 2023. **13**(1): p. 3637.

56. Chen, L.-C., et al., *Deeplab: Semantic image segmentation with deep convolutional nets, atrous convolution, and fully connected crfs.* IEEE transactions on pattern analysis and machine intelligence, 2017. **40**(4): p. 834-848.
57. Liu, S. and D. Huang. *Receptive field block net for accurate and fast object detection*. in *Proceedings of the European conference on computer vision (ECCV)*. 2018.
58. Kermany, D., *Labeled optical coherence tomography (oct) and chest x-ray images for classification.* Mendeley data, 2018.
59. Selvaraju, R.R., et al. *Grad-cam: Visual explanations from deep networks via gradient-based localization*. in *Proceedings of the IEEE international conference on computer vision*. 2017.
60. Jiang, P.-T., et al., *Layercam: Exploring hierarchical class activation maps for localization.* IEEE transactions on image processing, 2021. **30**: p. 5875-5888.